\documentclass[11pt]{article}
\usepackage[round]{natbib}
\ifdefined\usebigfont

\usepackage{times}
\usepackage[fontsize=13pt]{scrextend}
\usepackage[left=1.56in,right=1.56in,top=1.74in,bottom=1.74in]{geometry}
\else
\usepackage{fullpage}
\usepackage{times}
\fi

\usepackage{enumerate}
\usepackage{times}
\usepackage[OT1]{fontenc}
\usepackage{amsmath,amssymb}
\usepackage[usenames]{color}
\usepackage{multirow}
\usepackage[colorlinks,
            linkcolor=red,
            anchorcolor=blue,
            citecolor=blue
            ]{hyperref}

\usepackage[utf8]{inputenc} 
\usepackage[T1]{fontenc}    
\usepackage{booktabs}       
\usepackage{amsfonts}       
\usepackage{nicefrac}       
\usepackage{microtype}      

\usepackage{amsmath, amssymb}
\usepackage{algorithm}
\usepackage{algorithmic}
\usepackage{amsthm}
\usepackage[small]{caption}
\usepackage{graphicx}
\usepackage{subcaption}

\usepackage{enumitem}
\usepackage[olditem,oldenum]{paralist}

\newcommand{\norm}[1]{\left\lVert #1 \right\rVert}
\DeclareMathOperator{\EX}{\mathbb{E}}
\newtheorem{theorem}{Theorem}

\newtheorem{lemma}{Lemma}
\newtheorem{assumption}{Assumption}
\newtheorem{corollary}{Corollary}
\newcommand{\enqds}[1]{\mathbb{E} \left\lVert Q^{-1}#1 \right\rVert^2}
\newcommand{\ens}[1]{\mathbb{E} \left\lVert #1 \right\rVert^2}

\renewcommand{\S}{\mathcal{S}}
\newcommand{\A}{\mathcal{A}}
\newcommand{\T}{\mathcal{T}}

\newcommand{\B}{\mathcal{B}}
\newcommand{\Real}{\mathbb{R}}
\usepackage{hyperref}       

\hypersetup{
	plainpages=false,
	colorlinks=true,              
	linkcolor=blue,               
	anchorcolor=blue,             
	citecolor=blue,               
	filecolor=blue,               
	pagecolor=blue,               
	urlcolor=blue,                
	pdfview=FitH,                 
	pdfstartview=FitH,            
	pdfpagelayout=SinglePage      
}

\DeclareSymbolFont{extraup}{U}{zavm}{m}{n}
\DeclareMathSymbol{\varheart}{\mathalpha}{extraup}{86}
\DeclareMathSymbol{\vardiamond}{\mathalpha}{extraup}{87}

\title{\huge SVRG for Policy Evaluation with Fewer Gradient Evaluations}
\author{Zilun Peng $^\textbf{*}\,{}^\varheart$ \qquad Ahmed Touati $^\textbf{*}\,{}^\vardiamond \,{}^\clubsuit$ \qquad
Pascal Vincent ${}^\vardiamond \,{}^\clubsuit$
\qquad Doina Precup ${}^\varheart$
  }
\date{}

\begin{document}
\maketitle
\begin{abstract}
Stochastic variance-reduced gradient (SVRG) is an optimization method originally designed for tackling machine learning problems with a finite sum structure. SVRG was later shown to work for policy evaluation, a problem in reinforcement learning in which one aims to estimate the value function of a given policy. SVRG makes use of gradient estimates at two scales. At the slower scale, SVRG computes a full gradient over the whole dataset, which could lead to prohibitive computation costs. In this work, we show that two variants of SVRG for policy evaluation could significantly diminish the number of gradient calculations while preserving a linear convergence speed. More importantly, our theoretical result implies that one does not need to use the entire dataset in every epoch of SVRG when it is applied to policy evaluation with linear function approximation. Our experiments demonstrate large computational savings provided by the proposed methods.
\end{abstract}

\section{Introduction}

\let\thefootnote\relax\footnotetext{
$^\textbf{*}$ Equal contribution,
$^\vardiamond$ Mila, Universit\'e de Montr\'eal, 
$^\varheart$ Mila, McGill University, 
$^\clubsuit$ Facebook AI Research. \\ Short version of the paper is published in the proceedings of the 29th International Joint Conference on Artificial Intelligence and the 17th Pacific Rim International Conference on Artificial Intelligence (IJCAI-PRICAI2020)}

In reinforcement learning (RL), an agent continuously interacts with an environment by choosing actions as prescribed by a way of behaving called a policy. The agent observes its current state and performs an action based on its current policy (which is a probability distribution conditioned on state), then it reaches a new state and obtains a reward. The goal of the agent is to improve its policy, but a key requirement in this process is the ability to evaluate the expected long-term return of the current policy, called the value function. After evaluating the policy,  the policy can be updated so that more valuable states are visited more often. Performing policy evaluation efficiently is thus imperative to the success of training a RL agent.

Temporal difference (TD) learning \citep{sutton:td} is a classic method for policy evaluation, which uses the Bellman equation to bootstrap the estimation process and continually update the value function. The Least Squares Temporal Difference (LSTD) method \citep{lstd, boyan} is a more data-efficient approach which uses the data to construct a linear system approximating the original problem, then solves this system. It also has the advantage of not requiring a learning rate parameter. However, LSTD is not computationally feasible when the number of features $d$ is large, because it requires inverting a matrix of size $d\times d$.  When $d$ is large, stochastic gradient based approaches, such as GTD \citep{sutton:gtd}, GTD2 and TDC \citep{sutton:gtd2} are preferred because the amount of computation and storage during each update is linear in $d$. Compared to classical TD, these algorithms truly compute a gradient (instead of performing a fixed-point approximation which is in fact not a gradient update); as a result, they enjoy better theoretical guarantees, especially in the case of off-policy learning, in which the policy of interest for the evaluation is different from the policy generating the agent's experience.

Convex problems with large $n$ (number of data samples) and $d$ appear often in machine learning and there are many efficient stochastic gradient methods for finding solutions (e.g. SAG \citep{sag}, SVRG \citep{svrg}, SAGA \citep{saga}). In the problem of interests here, policy evaluation with linear function approximation, the objective function is a saddle-point formulation of the empirical Mean Squared Projected Bellman Error (MSPBE). It is convex-concave and not strongly convex in the primal variable, so existing powerful convex optimization methods do not directly apply. 

Despite this problem, \citet{du:stoc_var_pe} showed that SVRG and SAGA can be applied to solve the saddle point version of MSPBE with linear convergence rates, leading to fast, convergent methods for policy evaluation. An important and computationally heavy step of SVRG is to compute a full gradient at the beginning of every epoch. Subsequent stochastic gradient updates use this full gradient so that the variance of updating directions is reduced. In this paper, we address the computational bottleneck of SVRG by extending two methods, batching SVRG \citep{reza:practicalSVRG} and SCSG \citep{lei:scsg}, for policy evaluation. These methods were originally proposed to make SVRG computationally efficient when solving strongly convex problems, so they do not directly apply to our problem, a convex-concave function without strong convexity in the primal variable. 

In this work, we make the following key contributions:
\begin{compactenum}[\hspace{0pt} 1.]
    \setlength{\itemsep}{2pt}
    \item We show that both batching SVRG and SCSG achieve linear convergence rate for policy evaluation while yielding a  considerable saving in number of gradient computations. To the best of our knowledge, this is the first result for batching SVRG and SCSG in the saddle-point setting.
    \item While our analysis builds on the ideas of \citet{lei:scsg}, our proofs end up quite different and also a lot simpler because we exploit the structure of our problem. 
    \item  Our experimental results demonstrate that batching SVRG and SCSG are efficient in large data settings.
\end{compactenum}

\section{BACKGROUND}
In RL, a Markov Decision Process (MDP) is typically used to model the interaction between an agent and its environment. A MDP is defined by a tuple $(\S, \A, P, r,\gamma)$, where $\S$ is the set of possible states, $\A$ is the set of actions,the transition probability function $P : \S \times \A \rightarrow (\S \rightarrow [0,1])$ maps state-action pairs to distributions over next states. $r$ denotes the reward function: $(\S, \A) \rightarrow \rm \Real$, which returns the immediate reward that an agent will receive after performing an action $a$ at state $s$ and  $\gamma$ is the discount factor used to discount rewards received farther in the future. For simplicity, we will assume $\S$ and $\A$ are finite.

A policy $\pi : \S \rightarrow (\A \rightarrow [0, 1])$ is a mapping from states to distributions over actions.
The value function for policy $\pi$, denoted $V^{\pi}:\S \rightarrow \Real$, represents the expected sum of discounted rewards along the trajectories induced by the policy in the MDP: $V^{\pi}(s) = \EX [\sum_{t=0}^{\infty} \gamma^t r_t \mid s_0 = s, \pi]$.  $V^\pi$ can be obtained as the fixed point of the Bellman operator over the action-value function $\T^\pi V = r^\pi + \gamma P^\pi V$ where $r^\pi$ is the expected immediate reward and $P^\pi$ is defined as $P^\pi(s'|s) \triangleq  \sum_{a \in \A} \pi(a \mid s) P(s' \mid s, a)$.

In this paper, we are concerned with the policy evaluation problem~\citep{sutton1998introduction} i.e estimation of $V^\pi$ for a given policy $\pi$. In order to obtain generalization between different states, $V^\pi$ should be represented in a functional form. In this paper, we focus on linear function approximation of the form:
$V(s) \triangleq \theta^{\top}  \phi(s)$
where $\theta \in \Real^d$ is a weight vector and $\phi: \S \rightarrow \Real^d$ is a feature map from a states to a given $d$-dimensional feature space.

\section{OBJECTIVE FUNCTIONS}
We assume that the Markov chain induced by the policy $\pi$ is ergodic and admits a unique stationary distribution, denoted by $d^\pi$, over states. We write $D^\pi$ for the diagonal matrix whose diagonal entries are $(d^\pi(s))_{s \in \S}$.

If $\Phi = (\phi(s))_{s \in \S} \in \Real^{|\S| \times d}$ denotes the matrix obtained by stacking the state feature vectors row by row, then it is known \citep{bertsekas2011temporal} that $\Phi \theta^\star$ is the fixed point of the projected Bellman operator :
\begin{equation}
\Phi \theta^* = \Pi^{\pi} \T^\pi(\Phi \theta^*) \enspace ,
\label{eqn:bellman}
\end{equation}
where $\Pi^{\pi} = \Phi (\Phi^\top D^{\pi} \Phi)^{-1}\Phi^\top D^{\pi}$ is the orthogonal projection onto the space $S = \{ \Phi \theta \mid \theta \in \mathbb{R}^d \}$ with respect to the weighted Euclidean norm $||.||_{D^\pi}$. 
Rather than computing a sequence of iterates given by the projected Bellman operator, another approach for finding 
$\theta^*$ is to directly minimize \citep{sutton:gtd2,liu:gtd2} the Mean Squared Projected Bellman Error (MSPBE):
\begin{equation}
\mathbf{MSPBE}(\theta) = \frac{1}{2}||\Pi^{\pi} \T^\pi (\Phi \theta)- \Phi \theta ||^2_{D^\pi} \enspace .
\label{eqn:mspbe_vanilla}
\end{equation}

By substituting the definition of $\Pi^\pi$ into (\ref{eqn:mspbe_vanilla}), we can write MSPBE as a standard weighted least-squares problem (See \citet{sutton:gtd2} for a complete derivation):
\begin{align} \label{eqn:mspbe_abc}
    \mathbf{MSPBE}(\theta) = \frac{1}{2} \norm{A\theta - b}^2_{C^{-1}}
\end{align}
where $A$, $b$ and $C$ are defined as follows: $
    A = \mathbb{E}[\phi(s_t)(\phi(s_t)-\gamma \phi(s_{t+1}))^T] = \Phi^\top D^\pi(I - \gamma P^\pi) \Phi,
    b = \EX[\phi(s_t)r_t] = \Phi^\top D^\pi r^\pi \text{ and } C = \EX[\phi(s_t)\phi(s_t)^T] = \Phi^\top D^\pi \Phi, $
where the expectations are taken with respect to the stationary distribution.

\paragraph{Empirical MSPBE:} 
We focus here on the batch setting where we collect a dataset of $n$ transitions $\{(s_t, a_t, r_t, s_{t+1})\}_{t \in [n]}$ generated by the policy $\pi$.
We replace the quantities $A$, $b$ and $C$ in (\ref{eqn:mspbe_abc}) by their empirical estimates: 
\begin{equation}
    \hat A = \frac{1}{n} \sum_{t=1}^n \hat A_t, \quad \hat b = \frac{1}{n} \sum_{t=1}^n \hat b_t, \quad
    \hat C = \frac{1}{n} \sum_{t=1}^n \hat C_t . 
\end{equation}
where for all $t \in [n]$, for a given transition $(s_t, a_t, r_t, s_{t+1})$
\begin{align}
    \hat A_t  \triangleq \phi(s_t) (\phi(s_t) - \gamma \phi(s_{t+1}))^\top,
    \quad \hat b_t  \triangleq r_t \phi(s_t), 
    \quad \hat C_t \triangleq \phi(s_t) \phi(s_t)^\top .
\end{align}
Therefore we consider the empirical MSPBE defined as follows:
\begin{equation}\label{eqn:em-mspbe}
    \text{\textbf{EM-MSPBE}}(\theta) = \frac{1}{2} \norm{\hat A \theta - \hat b }^2_{\hat C^{-1}} .
\end{equation}

\paragraph{Finite sum structure:}
We aim at using stochastic variance-reduction techniques to our problem. These methods are designed for problem with finite sum structure as follows:
\begin{equation} \label{eq:min_finite_sum}
\min_x f(x) = \frac{1}{n}	\sum_{i=1}^{n} f_i(x) .
\end{equation}
Unfortunately, even by replacing quantities $A$, $b$ and $C$ by their finite-sample estimates, the obtained empirical objective in (\ref{eqn:em-mspbe}) could not be written in such form (\ref{eq:min_finite_sum}).
However, \cite{du:stoc_var_pe} convert the empirical MSPBE minimization in (\ref{eqn:em-mspbe}) into a convex-concave saddle point problem which presents a finite sum structure. To this end, \cite{du:stoc_var_pe} use the convex-conjugate trick. Recall that the convex conjugate of a real-valued function $f$ is defined as:
\begin{equation}
f^*(y)  = \sup_{x \in \mathit{X}} ( \langle y, x \rangle  - f(x))\enspace ,
\end{equation}
and $f$ is convex, we have $f^{**} = f$.
Also, if $f(x) = \frac{1}{2} ||x||^2_{\hat C^{-1}}$, then $f^{*}(x) = \frac{1}{2}||x||^2_{\hat C}$. 
Thanks to the latter relation, the empirical MSPBE minimization is equivalent to:
\begin{equation}\label{eq:saddle-point problem}
    \min_{\theta} \max_{\omega} \left( \langle \hat b - \hat A\theta, \omega \rangle - \frac{1}{2} || \omega||_{\hat C}^2 \right) .
\end{equation}
The obtained objective, we denote by $f(\theta, \omega)$, in (\ref{eq:saddle-point problem}) could be written as $f(\theta, \omega) = \frac{1}{n} \sum_{t=1}^n f_t(\theta, \omega)$ where $f_t(\theta, \omega) = \langle \hat b_t - \hat A_t\theta, \omega \rangle - \frac{1}{2} || \omega||_{\hat C_t}^2$

\section{EXISTING OPTIMIZATION ALGORITHMS}
Before presenting our new methods, we first review briefly existing algorithms that solve the saddle-point problem (\ref{eq:saddle-point problem}). Let's define the vector $F(\theta, \omega)$ obtained by stacking the primal and negative dual gradients:
\begin{equation} \label{eq:saddle-point gradient}
    F(\theta, \omega) \triangleq 
    \left (
    \begin{matrix}
    \nabla_\theta f(\theta, \omega) \\
    - \nabla_\omega f(\theta, \omega)
    \end{matrix}
    \right)
    = \left (
    \begin{matrix}
    - \hat A^\top \omega \\
    \hat A \theta - \hat b + \hat C \omega
    \end{matrix}
    \right) .
\end{equation}
We have $F(\theta, \omega) = \frac{1}{n}\sum_{t=1}^n F_t(\theta, \omega)$ where $\forall t \in [n]:$
$
    F_t(\theta, \omega) \triangleq
    \left (
    \begin{matrix}
    - \hat A_t^\top \omega \\
    \hat A_t \theta - \hat b_t + \hat C_t \omega
    \end{matrix}
    \right) $.

\paragraph{Gradient temporal difference:} GTD2 algorithm \cite{sutton:gtd2}, when applied to the batch setting, consists in the following update: for a uniformly sampled $t \in [n]$:

\begin{equation}
    \left( \begin{matrix} \theta \\ \omega \end{matrix} \right) \leftarrow 
    \left( \begin{matrix} \theta \\ \omega \end{matrix} \right) 
    - \left( \begin{matrix} \sigma_\theta & 0  \\ 0 & \sigma_\omega \end{matrix} \right) F_t(\theta, \omega) .
\end{equation}
where $\sigma_\theta$ and $\sigma_\omega$ are step sizes on $\theta$ and $\omega$. GTD2 has a low computation cost per iteration but only a sublinear convergence rate \citep{touati2018convergent}.

\paragraph{SVRG for policy evaluation:} \cite{du:stoc_var_pe} applied SVRG to solve the saddle-point problem (\ref{eq:saddle-point problem}). The idea is to alternate between full and stochastic gradient updates in two layers of loops. In the outer loop, a snapshot $(\tilde \theta, \tilde \omega)$ of the current variables is saved together
with its full gradients vector $F(\tilde \theta, \tilde \omega)$. Between snapshots, the variables are updated with a gradient
estimate corrected using the stochastic gradient:
\begin{equation}
    v_t \triangleq F_t(\theta, \omega) + F(\tilde \theta, \tilde \omega) - F_t(\tilde \theta, \tilde \omega) .
\end{equation}
where $t \in [n]$ is uniformly sampled.
\cite{du:stoc_var_pe} showed that the algorithm has a linear convergence rate although the objective (\ref{eq:saddle-point problem}) is not strongly convex in the primal variable $\theta$. However, the algorithm remains inefficient in term of computations as it requires to compute a full gradient using the entire dataset in the outer loop. In the rest of the paper, "An epoch" means an iteration of the outer loop.
In the sequel, we introduce two variants of SVRG for policy evaluation that alleviate the latter computational bottleneck while preserving the linear convergence rate.

\section{PROPOSED METHODS}

\subsection{Batching SVRG for Policy Evaluation}
Algorithm \ref{alg:batch_svrg} presents batching SVRG for policy evaluation. It applies batching SVRG \cite{reza:practicalSVRG} on solving the convex concave formulation of the empirical MSPBE.
\cite{reza:practicalSVRG} show that SVRG is robust to an inexact computation of the full gradient. In order to speed up the algorithm, we propose algorithm \ref{alg:batch_svrg}, similar to \cite{reza:practicalSVRG}, estimating the full gradient in each epoch $m$ using only a subset $\B_m$ (a mini-batch) of size $|\B_m| = B_m$ of training examples:
$\mu_m = \frac{1}{B_m} \sum_{t \in \B_m} F_t(\tilde{\theta}, \tilde{\omega})$
In each iteration $j$ of the inner loop in algorithm \ref{alg:batch_svrg}, it uses $v_{m,j}$ to update $\theta$ and $\omega$. $v_{m,j}$ is the usual SVRG update, except that the full gradients is replaced with the mini-batch gradients $\mu_m$:
    $v_{m,j} = F_{t_j}(\theta_{m,j}, \omega_{m, j}) - F_{t_j}(\tilde{\theta}, \tilde{\omega}) + \mu_m$
where $t_j$ is sampled uniformly in $[n]$.

\newsavebox{\algleft}
\newsavebox{\algright}
\savebox{\algleft}{%
\begin{minipage}{.49\textwidth}
\begin{algorithm}[H]\small
\caption{Batching SVRG for PE}
\label{alg:batch_svrg}
\textbf{Input}: initial point $(\theta, \omega)$, $\sigma_{\theta}, \sigma_{\omega}$, $M$ and $K$\\
\textbf{Output}: $(\theta, \omega)$
\begin{algorithmic}[1] 
\FOR{m = 0 to M-1}
\STATE Set $(\Tilde{\theta}, \Tilde{\omega})$ and $ (\theta_{m,0}, \omega_{m,0}) $ to $(\theta, \omega)$. 
\STATE \textcolor{red}{Choose a mini-batch size $B_m$}
\STATE Sample a set $\B_m$ with $B_m$ elements uniformly from $[n]$
\STATE \textcolor{red}{Compute $\mu_m = \frac{1}{|B_m|} \sum_{t \in \B_m} F_t(\tilde{\theta}, \tilde{\omega})$ }
\FOR{j = 0 to K-1}
\STATE Sample $t_j$ uniformly randomly from $[n]$
\STATE $v_{m,j} = F_{t_j}(\theta_{m,j}, \omega_{m, j}) - F_{t_j}(\tilde{\theta}, \tilde{\omega}) + \mu_m$
\STATE  \begin{align*}
\left( \begin{matrix}
           \theta_{m,j+1} \\
           \omega_{m,j+1}
         \end{matrix} \right) \leftarrow 
\left( \begin{matrix}
           \theta_{m,j} \\
           \omega_{m,j}
         \end{matrix} \right) - 
\left( \begin{matrix}
           \sigma_{\theta} & 0\\
           0 & \sigma_{\omega}
         \end{matrix} \right)
         v_{m, j}      
\end{align*}
\ENDFOR
\STATE $(\theta, \omega) = (\theta_{m,K}, \omega_{m,K})$
\ENDFOR
\STATE \textbf{return} $(\theta, \omega)$
\end{algorithmic}
\end{algorithm}
\end{minipage}
}%
\savebox{\algright}{%
\begin{minipage}{.49\textwidth}
\begin{algorithm}[H] \small
\caption{SCSG for PE}
\label{alg:scsg}\textbf{Input}: initial point $(\theta, \omega)$, $\sigma_{\theta}, \sigma_{\omega}$, $M$, $K$ and $B$\\
\textbf{Output}: $(\theta, \omega)$
\begin{algorithmic}[1] 
\FOR{m = 1 to M}
\STATE Set $(\Tilde{\theta}, \Tilde{\omega})$ and $ (\theta_{m,0}, \omega_{m,0}) $ to $(\theta, \omega)$. 
\STATE Sample a set $\B$ with $B$ elements uniformly from $[n]$
\STATE \textcolor{red}{Compute $\mu_m = \frac{1}{B} \sum_{t \in \B} F_t(\tilde{\theta}, \tilde{\omega})$}
\STATE \textcolor{red}{$K_m \sim \text{Geom}(\frac{B}{B+1})$} 
\FOR{j = 0 to $K_m-1$}
\STATE Sample $t_j$ uniformly randomly from $[n]$
\STATE $v_{m,j} = F_{t_j}(\theta_{m,j}, \omega_{m, j}) - F_{t_j}(\tilde{\theta}, \tilde{\omega}) + \mu_m$
\STATE  \begin{align*}
\left( \begin{matrix}
           \theta_{m,j+1} \\
           \omega_{m,j+1}
         \end{matrix} \right) \leftarrow 
\left( \begin{matrix}
           \theta_{m,j} \\
           \omega_{m,j}
         \end{matrix} \right) - 
\left( \begin{matrix}
           \sigma_{\theta} & 0\\
           0 & \sigma_{\omega}
         \end{matrix} \right)
         v_{m, j}      
\end{align*}
\ENDFOR
\STATE $(\theta, \omega) = (\theta_{m,K_m}, \omega_{m,K_m})$
\ENDFOR
\STATE \textbf{return} $(\theta, \omega)$
\end{algorithmic}
\end{algorithm}
\end{minipage}
}%
\noindent\usebox{\algleft}\hfill\usebox{\algright}%

\subsection{Stochastically Controlled
Stochastic Gradient (SCSG) for Policy Evaluation}
Algorithm \ref{alg:scsg} presents Stochastically Controlled Stochastic Gradient (SCSG) for Policy Evaluation. SCSG is initially introduced for convex minimization in \citet{lei:scsg}. Here, we apply it to our convex-concave saddle-point problem. Similar to Bachting SVRG for policy evaluation in algorithm \ref{alg:batch_svrg}, algorithm \ref{alg:scsg} implements the gradient computation on a subset $\B$ of training examples at each epoch, but the mini-batch size is fixed in advance and not varying. Moreover, instead of being fixed, the number of iteration for the inner loop in algorithm \ref{alg:scsg} is sampled from a geometrically
distributed random variable: $K_m \sim \text{Geom}(\frac{B}{B+1})$ for each epoch $m$.

\section{CONVERGENCE ANALYSIS}
\subsection{Notations and Preliminary}
In order to characterize the convergence rates of the proposed algorithms \ref{alg:batch_svrg} and \ref{alg:scsg}, we need to introduce some new notations and state new assumptions.

We denote by $\|A\| \triangleq \sup_{\|x\|=1}\| Ax\|$ the spectral norm of the matrix A and by $\kappa(A) = \|A\| \| A^{-1}\|$ its condition number. If the eigenvalues of a matrix $A$ are real, we use $\lambda_{\max}(A)$ and $\lambda_{\min}(A)$ to denote respectively the largest and the smallest eigenvalue.

If we set $\sigma_\omega = \beta \sigma_\theta$ for a positive constant $\beta$, it is possible to write the inner loop update (line 9 in both algorithms) as an update for the vector $z_{m,j} \triangleq  \left( \begin{matrix}
\theta_{m,j} \\
\frac{1}{\sqrt{\beta}}\omega_{m,j}
\end{matrix} \right) $ as follows : 
\begin{equation*}
    z_{m,j+1} = z_{m,j} - \sigma_\theta \left(G_{t_j} z_{m,j} + (G_m - G_{t_j}) z_{m, 0} - g_m \right)
\end{equation*}
where:
\begin{equation*}
G_t \triangleq \left( \begin{matrix}
0 & - \sqrt{\beta} \hat{A}_t^\top \\
\sqrt{\beta} \hat{A}_t & \beta \hat{C}_t
\end{matrix}
\right),
\quad 
g_t \triangleq  \left( \begin{matrix}
0 \\
\sqrt{\beta} \hat{b}_t
\end{matrix} \right) 
\end{equation*}
and their corresponding averages over the mini-batch $B_m$:
\begin{equation*}
G_m \triangleq \frac{1}{B_m} \sum_{t \in \B_m} G_t, \quad 
g_m \triangleq \frac{1}{B_m} \sum_{t \in \B_m} g_t
\end{equation*}
Let's now define the matrix $G$ (the vector $g$) as the average of matrices $G_t$ (vectors $g_t$) over the entire dataset:
\begin{equation*}
G \triangleq \left( \begin{matrix}
0 & -\sqrt{\beta} \hat A^\top \\
\sqrt{\beta} \hat A & \beta \hat C
\end{matrix} \right) 
\quad \text{and} \quad 
g \triangleq \left( \begin{matrix}
0 \\ \sqrt{\beta} \hat b
\end{matrix} \right)
\end{equation*}
To simplify notations, we overload the notation $\lambda_{\min} = \lambda_{\min}(G)$.
Another important quantity that characterizes  \textit{smoothness} of our problem is $L_G^2$ defined below as:

\begin{equation}
\label{defn:conv_anal_LG}
    L_{G}^2 = \norm{\frac{1}{n} \sum_{t=1}^{n} G_t^\top G_t} 
\end{equation}
The matrix $G$ will play a key role in the convergence analysis of both algorithms \ref{alg:batch_svrg} and \ref{alg:scsg}. \cite{du:stoc_var_pe} have already studied the spectral properties of $G$ as it was critical for the convergence of SVRG for policy evaluation. The following lemma, restated from \citep{du:stoc_var_pe}, show the condition $\beta$ should satisfy so that $G$ is diagonalizable with all its eigenvalues real
and positive.

\begin{assumption} \label{assump:non-singularity} $\hat A$ nonsingular and $\hat C$ is definite positive. This implies that the saddle-point problem admits a unique solution $(\theta^{\star}, \omega^{\star}) = (\hat A^{-1} \hat b, 0)$ and we define $z^{\star} \triangleq (\theta^{\star}, \frac{1}{\sqrt{\beta}}\omega^{\star})$. 
\end{assumption}

\begin{lemma} \citep{du:stoc_var_pe} \label{lem:G_diag_cond}
Suppose assumption \ref{assump:non-singularity} holds and if we choose $\beta = \frac{\sigma_\omega}{\sigma_\theta} =  \frac{8\lambda_{\max}(\hat{A}^T \hat{C}^{-1} \hat{A})}{\lambda_{\min}(\hat{C})}$, then the matrix $G$ is diagonalizable with all its eigenvalues real and positive.
\end{lemma}

If assumptions of lemma \ref{lem:G_diag_cond} hold, we can write $G$ as $G = Q \Lambda Q^{-1} $ where $\Lambda$ is a diagonal matrix whose diagonal entries are the eigenvalues of $G$, and $Q$ consists of it eigenvectors as column. We define the residual vector $\Delta_m = z_m - z^\star$. To study the behaviour of our algorithms, we use the potential function $\norm{Q^{-1}\Delta_m}^2$. As $\norm{Q}^2 \norm{Q^{-1}\Delta_m}^2 \geq \norm{\Delta_m}^2 \geq \norm{\theta - \theta^\star}^2$, the convergence of $\norm{Q^{-1}\Delta_m}^2$ implies the convergence of $\norm{\theta - \theta^\star}^2$.

\subsection{Convergence of batching SVRG for Policy Evaluation}
In order to study the behavior of algorithm \ref{alg:batch_svrg}, we defined $e_m$ the error occurred at epoch $m$. This error comes from computing  the gradients over a mini-batch $\B_m$ instead of the entire dataset.
\begin{equation}\label{eq: error def}
    e_m = (G z_m - g) - (G_m z_m - g_m)
\end{equation}
The stochastic update of the inner loop could be written as follows:
\begin{equation}
    z_{m,j+1} = z_{m,j} - \sigma_\theta \left(G_{t_j} z_{m,j} + (G - G_{t_j}) z_{m} - g - e_m \right)
\end{equation}
\begin{theorem} \label{prop:one_epoch_anal}
Suppose assumption \ref{assump:non-singularity} holds and if we choose $\sigma_\theta = \frac{\lambda_{\min}}{6 \kappa(Q)^2 L_G^2}$, $\beta = \frac{8\lambda_{\max}(\hat{A}^T \hat{C}^{-1} \hat{A})}{\lambda_{\min}(\hat{C})}, \sigma_\omega = \beta \sigma_\theta$ and $K = \frac{2}{\sigma_\theta \lambda_{\min}} = 
    \frac{12 \kappa(Q)^2 L_G^2}{\lambda_{\min}^2}$,
then we obtain:

\begin{align} \label{eq:batch_ineq}
    \enqds{\Delta_{m+1}}
&\leq \underbrace{ \frac{2}{3} \EX\norm{Q^{-1}\Delta_{m}}^2}_{\text{linear convergence term}} + \underbrace{2  \frac{1 - \sigma_\theta\lambda_{\min} }{ \lambda_{\min}^2} \EX \norm{Q^{-1}e_m}^2}_{\text{additional error term}}
\end{align}
\end{theorem}

\begin{proof}
See appendix \ref{appen:bsvrg_proof}
\end{proof}

Note that if $B_m=n \quad \forall m$, the error is zero and we recover the convergence rate of SVRG in theorem \ref{prop:one_epoch_anal}. Moreover, we could still maintain the linear convergence rate if the error term vanishes at an appropriate rate.
In particular, the corollary below provides a possible batching strategy to control the error term.
\begin{corollary} \label{coroll: batch_svrg}
Suppose that assumptions of theorem \ref{prop:one_epoch_anal} hold. If the sample variance of the norms of the vectors $F_t$ is bounded by a constant $\Xi^2$:
    $\frac{1}{n-1} \sum_{t=1}^n \norm{G_t z - g_t}^2 - \norm{G z - g}^2 \leq \Xi^2 \quad \forall z$
and we set $B_m = \frac{n \Xi^2}{\Xi^2 + n \alpha \rho^m} < n$ for some constants $\alpha > 0$ and $\rho<2/3$ then we obtain:
\begin{align}
    \enqds{\Delta_{M}}
 &\leq \left(\frac{2}{3} \right)^M  \bigg(\enqds{\Delta_{0}} + \frac{3 \alpha( 1 - \sigma_\theta\lambda_{\min}) }{ \lambda_{\min}^2 (1 - 3\rho/2)} \| Q^{-1}\|^2  \bigg)
\end{align}
\end{corollary}

Proof of corollary \ref{coroll: batch_svrg} is in appendix \ref{appendix: bsvrg_corollary}. We conclude that an exponentially-increasing schedule of mini-batch sizes achieves linear convergence rate for batching SVRG. This batching strategy saves many gradients computations in early stages of the algorithm comparing to vanilla SVRG.

\subsection{Convergence of SCSG for Policy Evaluation}
Algorithm \ref{alg:scsg} considers a fixed mini-batch size $B$ instead of varying size as in algorithm \ref{alg:batch_svrg}. Moreover, the number of iteration $K_m$ of the inner loop is sampled from a geometric distribution, i.e. $K_m \sim \text{Geom}(\frac{B}{B+1})$, which implies that the number of iteration is equal in expectation to $\frac{\frac{B}{B+1}}{1 - \frac{B}{B+1}} = B$. 

Before stating the convergence result, we introduce \textit{the complexity measure} defined as follows:
\begin{equation} \label{eq: complexity measure}
    \mathcal{H} = \frac{1}{n} \sum_{t=1}^n \| G_t z^\star - g_t \|^2
\end{equation}
This quantity is equivalent to the complexity measure that is introduced by \cite{lei:scsg} to motivate and analyze SCSG for convex finite sum minimization problem, $\min_{x} f(x) = \frac{1}{n} \sum_{i}^n f_i(x)$ and that is defined as:
\begin{equation}
    \mathcal{H}(f) = \inf_{x^\star \in \arg\min f(x)} \frac{1}{n}\sum_{i=1}^n \| \nabla f_i(x^*) \|^2
\end{equation}

\begin{theorem} \label{thm:scsg_result}
Suppose assumption \ref{assump:non-singularity} holds. Set $\sigma_\theta \leq \min\{\frac{\lambda_{min}}{20\kappa(Q)^2 L_G^2}, \frac{5}{28B\lambda_{max}}\}$, $\sigma_\omega = \beta \sigma_\theta$ and $B \geq \frac{70\kappa(Q)^2 L_G^2}{\lambda_{min}^2}$. The last iterate $\theta_{M-1}$ satisfies: 

\begin{align}
\EX\norm{\theta_{M-1} - \theta^\star}^2 &\leq \frac{(1+0.7\sigma_\theta B\lambda_{max})\kappa(Q)^2}{(1+0.8\sigma_\theta B\lambda_{min})^M}\ens{\Delta_0} + \frac{60\kappa(Q)^2 \mathcal{H} I(B<n)}{B\lambda_{min}^2}
\end{align}
\end{theorem}

\textit{Proof sketch: }
After taking the expectation, squared two norms and some manipulations on SCSG's update, we have:
\begin{align} \label{proof_sketch:scsg_update}
    2\sigma_\theta \EX \left<\Lambda Q^{-1}\Delta_{m, K_m}, Q^{-1}\Delta_{m, K_m} \right>  & \leq \frac{1}{B}\left(\enqds{\Delta_{m,0}} - \enqds{\Delta_{m, K_m}} \right) \nonumber\\
    \quad &+ 2\sigma_\theta \EX \left<Q^{-1}e_m, Q^{-1}\Delta_{m, K_m} \right> \nonumber\\
    \quad &+2 \sigma_\theta^2\kappa(Q)^2 L_G^2 \enqds{\Delta_{m,K_m}} \nonumber\\ 
    \quad &+ 2\sigma_\theta^2 \kappa(Q)^2 L_G^2\enqds{\Delta_{m}} \nonumber\\
    \quad &+ 2\sigma_\theta^2\EX\norm{\Lambda Q^{-1}\Delta_{m,K_m}}^2 + 2\sigma_\theta^2\norm{Q^{-1}e_m}^2
\end{align}

We do not make smoothness assumptions on individual gradient functions as \cite{lei:scsg} did, so we have to deal with $\enqds{\Delta_{m,K_m}}$ and $\enqds{\Delta_{m}}$, which are terms resulting from variances of SCSG's update. By taking the advantage that $\Lambda$ is a diagonal matrix consisting of positive eigenvalues, we have $\enqds{x} \leq \frac{1}{\lambda_{min}}\EX x^\top Q^{-\top} \Lambda Q^{-1}x$. This helps us to manipulate terms in (\ref{proof_sketch:scsg_update}). Note that by assumption \ref{assump:non-singularity}, lemma \ref{lem:G_diag_cond} ensures $G=Q\Lambda Q^{-1}$ where $\Lambda$ contains eigenvalues of $G$ that are real and positive. For the complete proof, see appendix \ref{appendix: scsg_proof}.

\begin{corollary} \label{cor:scsg_comp_cost}
Suppose assumption \ref{assump:non-singularity} holds. Set $\sigma_\theta = \min\left\{\frac{\lambda_{min}}{20\kappa(Q)^2 L_G^2}, \frac{5}{28B\lambda_{max}}\right\}$, $\sigma_\omega = \beta \sigma_\theta$ and $B = \min\left\{\max\left\{\frac{70\kappa(Q)^2 L_G^2}{\lambda_{min}^2}, \frac{120\kappa(Q)^2\mathcal{H}}{\lambda_{min}^2 \epsilon}\right\}, n\right\}$. Let $\epsilon > 0$, the computational cost in expectations that algorithm \ref{alg:scsg} required to obtain $\EX \norm{\theta_{M-1} - \theta^\star}^2 \leq \epsilon$ is: 

\begin{equation*}
    \mathcal{O}\left(\left(\min\left\{ \frac{\kappa(Q)^2\mathcal{H}}{\lambda_{min}^2 \epsilon}, n\right\} + \frac{\kappa(Q)^2 L_G^2}{\lambda_{min}^2} \right)d\times \log\left(\frac{\kappa(Q)^2\ens{\Delta_0}}{\epsilon} \right)\right)
\end{equation*}
\end{corollary}

See appendix \ref{appendix: scsg_corollary} for the proof of corollary \ref{cor:scsg_comp_cost}. Table \ref{table: rates} lists sample complexities of our methods and other related methods. GTD2 is the cheapest computationally but it has a sublinear convergence rate. Both SVRG and SCSG achieve linear convergence rates. When the dataset size $n$ is small, SVRG and SCSG have an equivalent computational cost. However, when $n$ is large and the required
accuracy $\epsilon$ is low, SCSG saves unnecessary computations and is able to achieve the
target accuracy with potentially less than a single pass through the dataset. 

\begin{table}[tbh]
 \caption{
 \label{table: rates} 
  Computational costs of different gradient based policy evaluation algorithms. We report the computational cost of GTD2 from \citep{touati2018convergent} and computational costs of SVRG and SAGA from \citep{du:stoc_var_pe}. We use quantities in our result to represent their computational costs. 
 }
\begin{center}
\begin{tabular}{l l}
 \hline
 Algo & Computational Cost \\
 \hline
 GTD2 & $\mathcal{O}\left(\frac{ \kappa(Q)^2\mathcal{H}d}{\lambda_{\min}^2 \epsilon}\right)$\\
 
 SAGA & $\mathcal{O}\left(\left(n  + \frac{\kappa(Q)^2 L^2_G}{ \lambda_{\min}^2 }\right)d\times \ln(1/\epsilon)\right)$ \\
  
 SVRG & $\mathcal{O}\left(\left(n  + \frac{\kappa(Q)^2 L^2_G}{ \lambda_{\min}^2 }\right)d\times \ln(1/\epsilon)\right)$ \\
 
Batching SVRG & $\mathcal{O}\left(\left(n'  + \frac{\kappa(Q)^2 L^2_G}{ \lambda_{\min}^2 }\right)d\times \ln(1/\epsilon)\right)$ \\
 
 SCSG & $\mathcal{O}\left(\left(\frac{\kappa(Q)^2 \mathcal{H}}{\lambda_{\min}^2 \epsilon} \land n  + \frac{\kappa(Q)^2 L^2_G}{ \lambda_{\min}^2 }\right)d\times \ln(1/\epsilon)\right)$ \\
 \hline
 \end{tabular}
 \end{center}
 \end{table}
 
 We compare the computational costs of batching SVRG and SVRG. In epoch $m$, batching SVRG's computational cost is $O(B_m + K_m)$, where $B_m$ is the batch size and $K_m$ is the number of inner loop iteration. According to theorem \ref{prop:one_epoch_anal}, $K_m$ is set as $O(\frac{\kappa(Q)^2 L_G^2}{\lambda_{\min}^2})$ for all $m$. If we use an exponentially increasing sequence of batch sizes that we considered in corollary \ref{coroll: batch_svrg}, $B_m$ is strictly less than $n$, for all $m$. Since batching SVRG converges linearly, it takes $O(\ln(1/\epsilon))$ epochs to reach an $\epsilon$-optimal solution. The overall computational cost of batching SVRG is $O\left(\left(n'  + \frac{\kappa(Q)^2 L^2_G}{ \lambda_{\min}^2 }\right)d\times \ln(1/\epsilon) \right)$ as shown in table \ref{table: rates}, where $n' < n$, so batching SVRG is computationally more efficient than the vanilla SVRG.
 
 \section{RELATED WORKS}
Stochastic gradient methods \citep{sgd} is the most popular method for optimizing convex problems with a finite sum structure, but has slow
convergence rate due to the inherent variance. Later, various works show that a faster convergence rate is possible provided that the objective function is strongly convex and smooth. Some representative ones are SAG, SVRG, SAGA \citep{sag,svrg,saga}. Among these methods, SVRG has low memory requirements but requires a lot of computations. There have been attempts to make SVRG computationally efficient for minimizing convex problems \citep{reza:practicalSVRG,lei:scsg}, but they do not directly apply to the problem of our interests, a convex-concave saddle-point problem without strong convexity in the primal variable. A general convex-concave saddle-point problem can be solved with linear convergence rate \citep{balamurugan&bach}, but their method requires strong convexity in the primal variable and the proximal mappings of variables in our problem are difficult to compute \citep{du:stoc_var_pe}.

Many existing works study policy evaluation with linear function approximation. Gradient based approaches \citep{rg_baird,sutton:gtd,sutton:gtd2,liu:gtd2} choose different objective functions and parameters of the value function are optimized toward solutions of their objective functions. Least square approaches \citep{lstd,boyan} directly compute the closed form solutions and have high computation costs because they need to compute matrix inverses. The idea of SVRG has been applied to policy evaluation \cite{ctd,du:stoc_var_pe}. In this work, we extend SVRG for policy evaluation, proposed in \citet{du:stoc_var_pe},   and show that the amount of computations can be reduced with linear convergence guarantees.

In control case, \citet{papini2018stochastic} adapt SVRG to policy gradient and they use mini-batch to approximate the full-gradient similarly to our work. However, their problem is a non-convex minimization and they obtain a sublinear convergence rate. 

\begin{figure*}[hbt!]
\centering
\begin{tabular}{cccc}
    \includegraphics[width=0.22\linewidth]{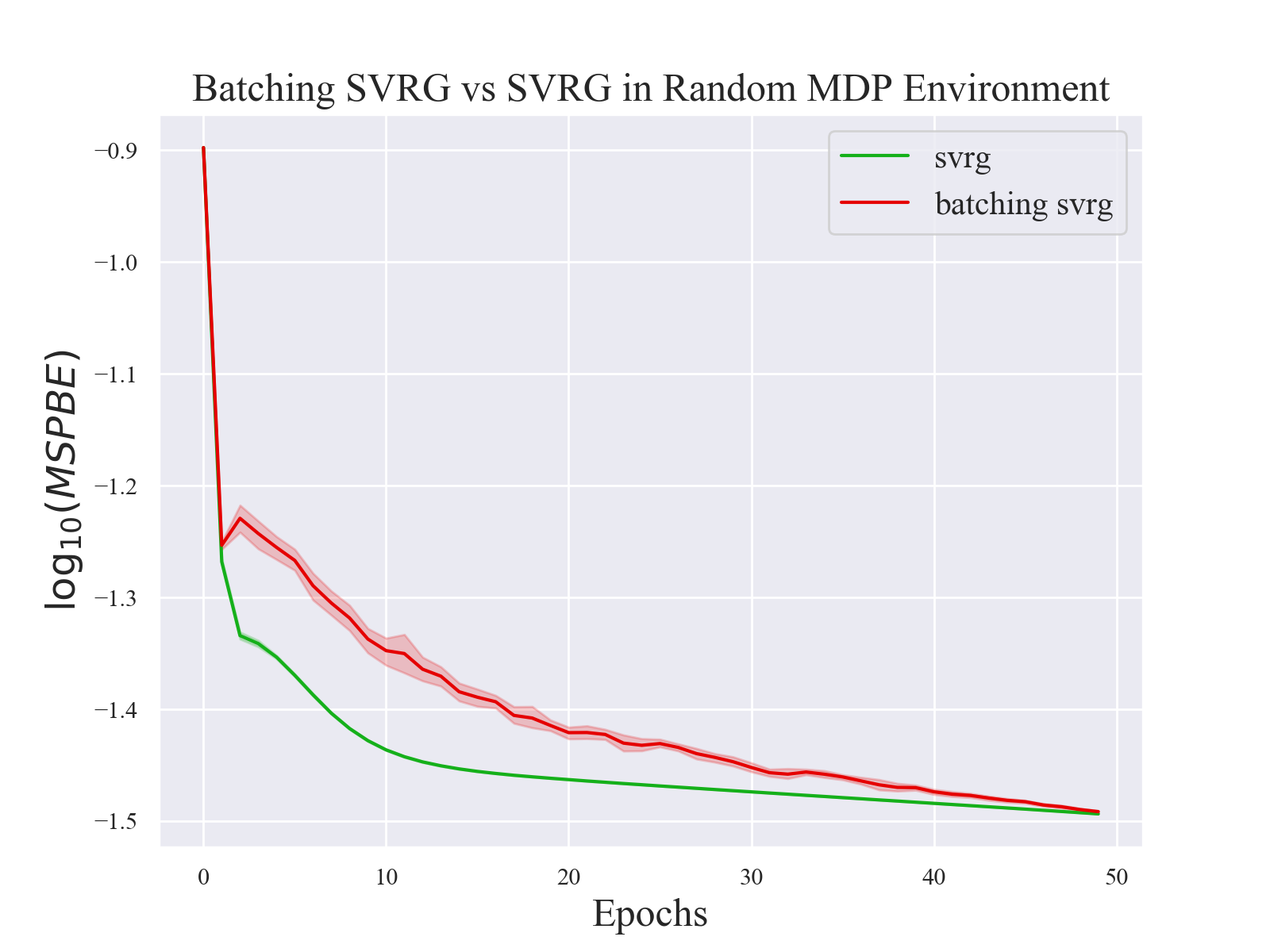}&
    \includegraphics[width=0.22\linewidth]{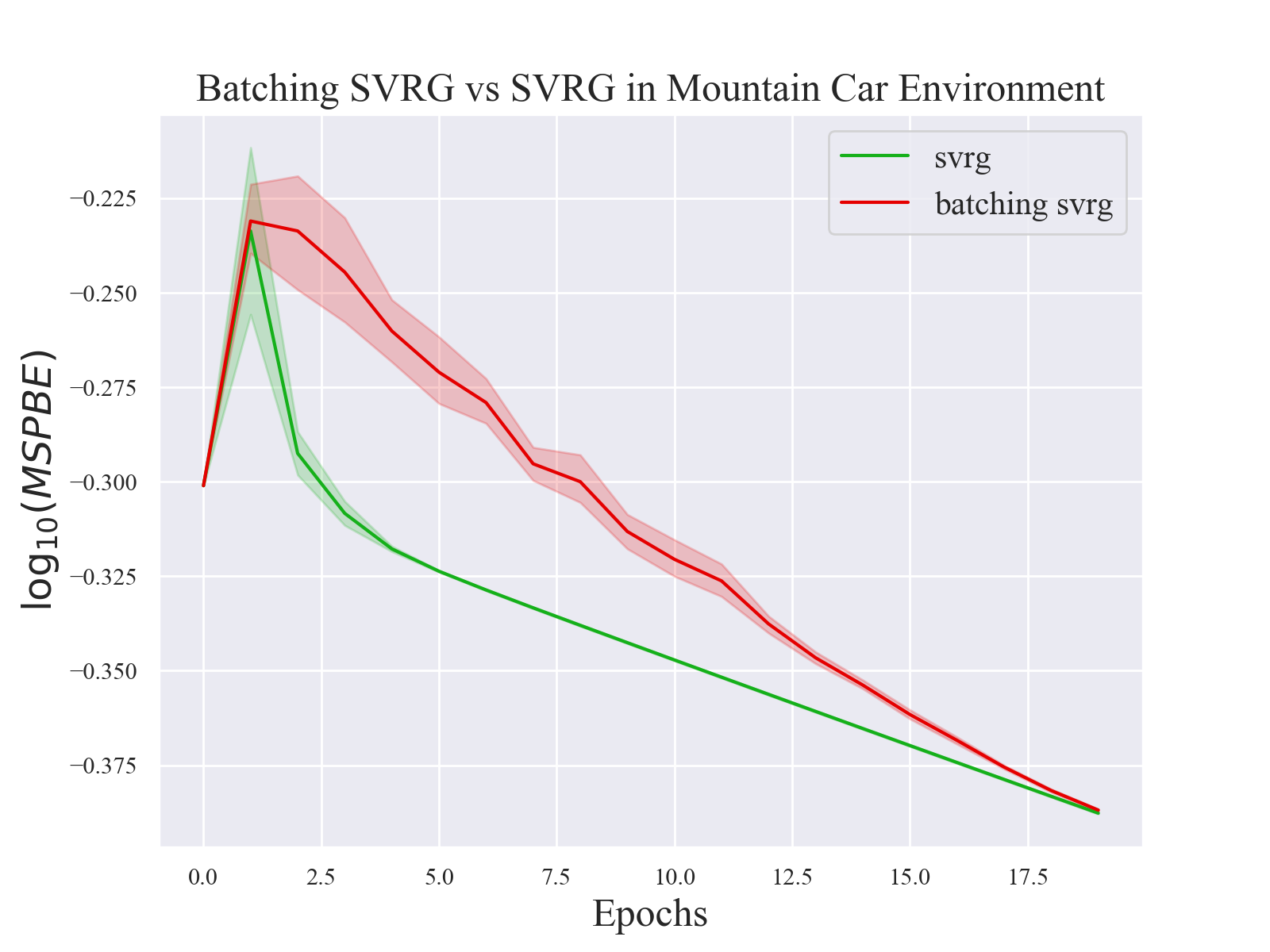} &
    \includegraphics[width=0.22\linewidth]{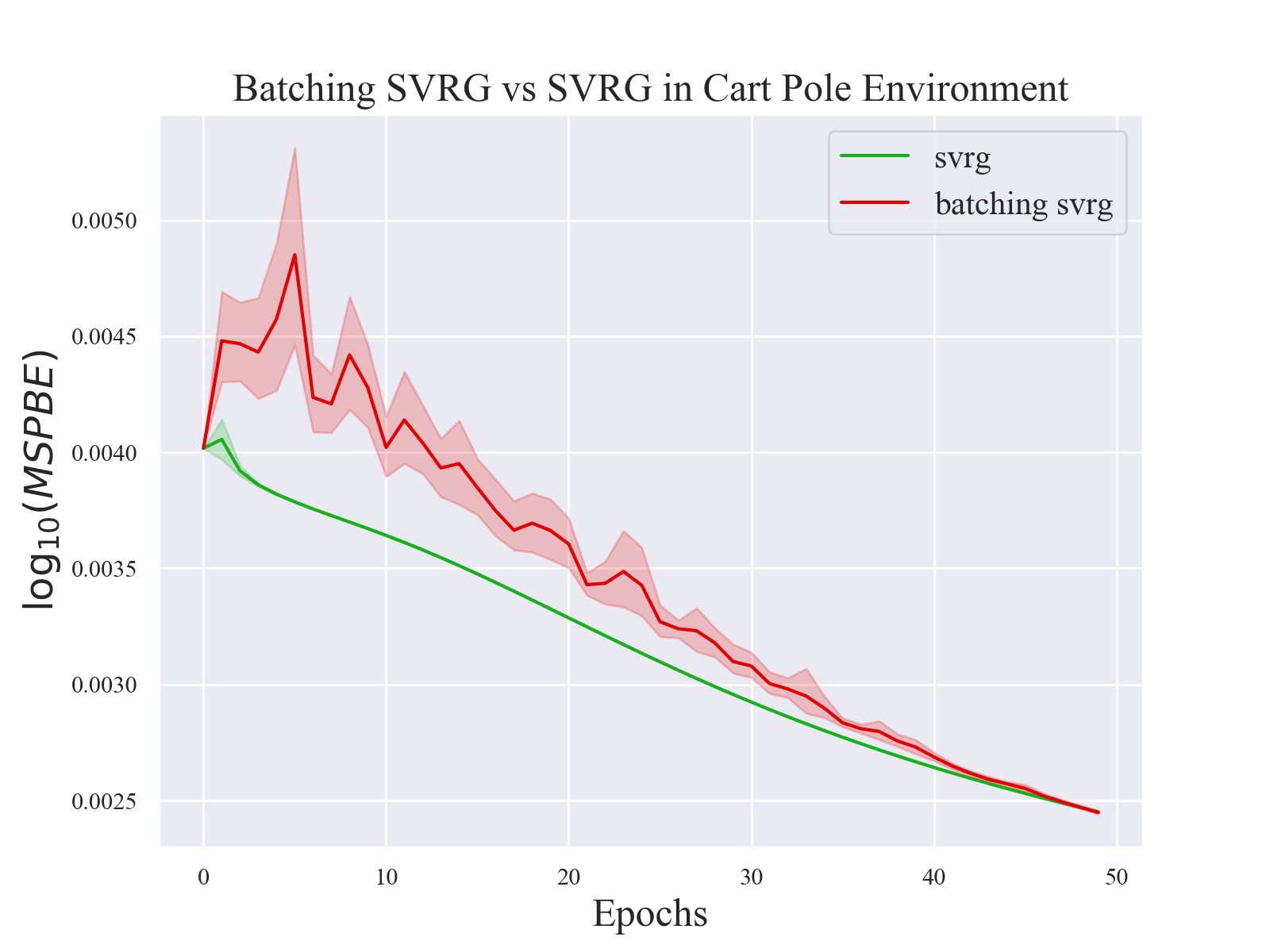}& 
    \includegraphics[width=0.22\linewidth]{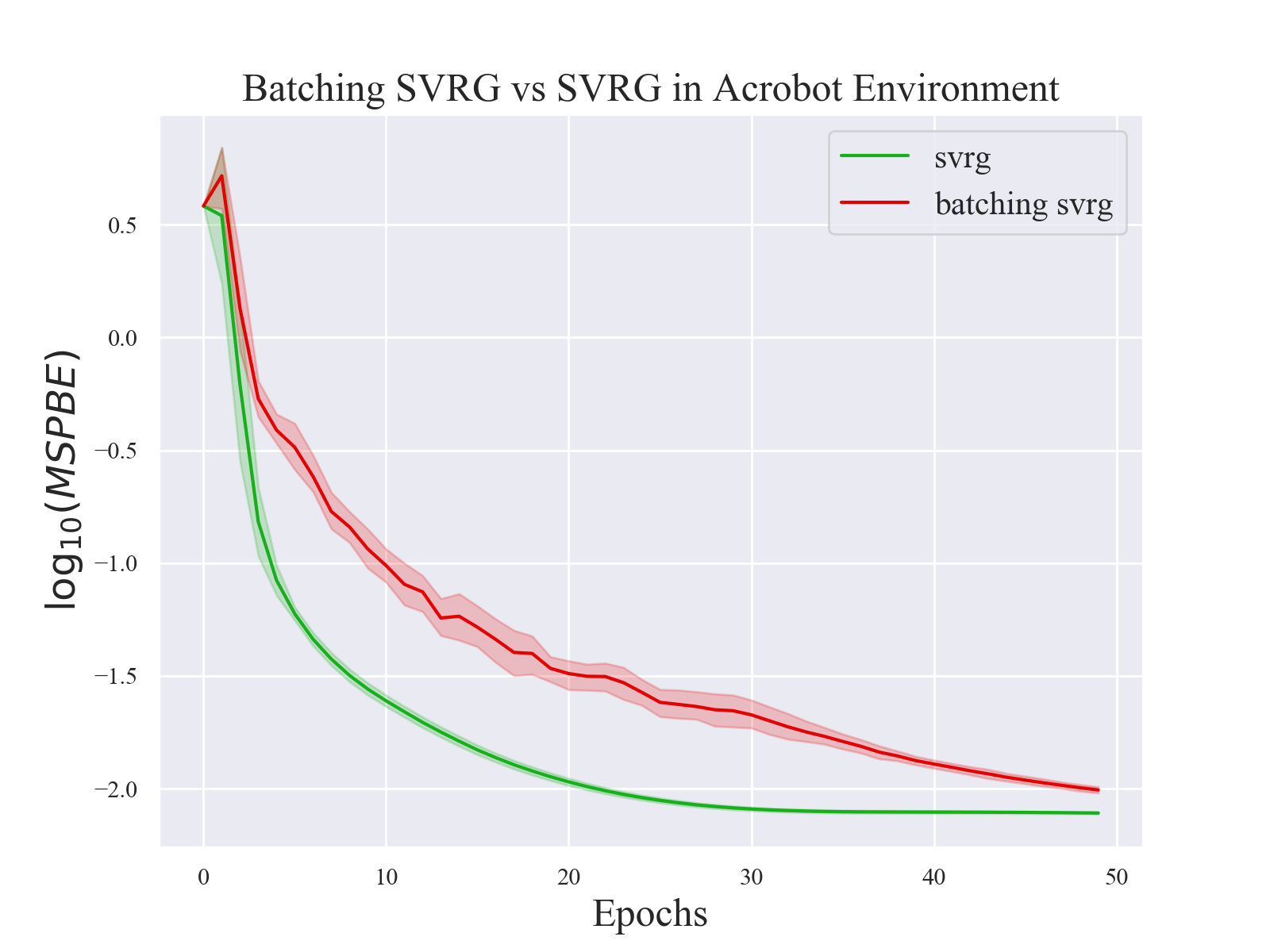}
\end{tabular}
\caption{Policy evaluation performances of batching SVRG and SVRG in different environments. Note that batching SVRG and SVRG were run for same number of epochs, and batching SVRG achives same level of performances with SVRG while taking fewer passes through the dataset (shown in table \ref{tab:comp_cost_bsvrg_svrg}).}
\label{fig:batch_svrg_vs_svrg}
\end{figure*}

\section{EXPERIMENTS}
Our proposed methods, batching SVRG and SCSG, are evaluated with LSTD, SVRG, SAGA and GTD2 on 4 tasks: Random MDP \citep{dann:survey}, MountainCar-v0, CartPole-v1 and Acrobot-v1 \citep{openai_gym}. More details on experiment setup and procedures can be found in appendix \ref{appen:exp_detail}. 

\subsection{Comparing batching SVRG and SVRG}

We show empirically that batching SVRG converges as fast as SVRG while using less amount of data. Figure \ref{fig:batch_svrg_vs_svrg} shows policy evaluation results of SVRG and batching SVRG in different environments, and table \ref{tab:comp_cost_bsvrg_svrg} shows computational costs of SVRG and batching SVRG. Given same number of epochs, batching SVRG achieves same level of performances with SVRG while taking fewer passes through the dataset. We observe that batching SVRG's performances are not stable and are worse than SVRG at the beginning, but objective values keep decreasing. This means that full gradients are estimated reasonably well by batch gradients in early epochs. As batching SVRG starts to use full gradients in later epochs, it reaches a same level of performances with SVRG. The empirical performance of batching SVRG is expected because our theoretical result suggests that having an approximation error of the full gradient will not affect the overall convergence speed if the error decreases properly. 

\begin{table}[hbt!]
    \centering
    \caption{Computational costs of batching SVRG and SVRG. This table shows the number of passes through the dataset for SVRG and batching SVRG when generating results given in Figure \ref{fig:batch_svrg_vs_svrg}.}
    \begin{tabular}{|c|c|c|}
    \hline
     Tasks    & SVRG & Batching SVRG\\ \hline
    Random MDP & 100 & \textbf{71}\\ \hline
    Mountain Car & 40 & \textbf{31}\\ \hline
    Cart Pole & 100 & \textbf{71}\\ \hline
    Acrobot & 100 & \textbf{71}\\ \hline
    \end{tabular}
    \label{tab:comp_cost_bsvrg_svrg}
\end{table}

\begin{table*}[]
\caption{All methods' performances in control. All values in the table are number of steps each method takes to reach the terminal state. In \textit{Mountain Car} and \textit{Acrobot}, small values mean good performances in control. In \textit{Cart pole}, large values mean good performances in control. First column shows names of all tasks.}

    \resizebox{\textwidth}{!}{
    \begin{tabular}{|c|c|c|c|c|c|c|}
    \hline
     Task         & GTD2 & SVRG & SAGA & Batching SVRG & SCSG & LSTD \\ \hline
  Mountain Car & $348\pm181$ &$186\pm149$  &$\mathbf{166\pm125}$ &$170\pm135 $ &$276\pm178$ & $292\pm187 $\\ \hline
  Cart Pole  & $163\pm95$ & $\mathbf{280\pm88}$ & $246\pm75$ & $\mathbf{283\pm82}$ & $217\pm100$ & $183\pm80 $ \\ \hline
  Acrobot & $431\pm128$ & $\mathbf{96\pm6}$ & $101\pm6$ & $\mathbf{97\pm6}$ & $126\pm78$ & $176\pm157 $   \\ \hline
  Mountain Car (large data) & $197\pm12$ & $200\pm0 $  & $200\pm0 $ &$200\pm0 $ &$199\pm7 $ & $\mathbf{116\pm6}$\\ \hline
  Cart Pole (large data) & $155\pm84$ & 9 & 9 & $297\pm100$ & $258\pm111$ & $\mathbf{290\pm26}$ \\ \hline
  Acrobot (large data) & $445\pm96$ & 500 & 500 & $113\pm26$ & $97\pm6$ & $\mathbf{91\pm4}$   \\ \hline
    \end{tabular}}
    \label{tab:control_results}
\end{table*}

\subsection{Control performances}

We run all gradient based methods and LSTD for policy evaluation and apply the learned policy on control tasks. This lets us test the practicality of our methods. We also intend to test our proposed methods' performances in large datasets. In \textit{Mountain Car (large data), Cart Pole (large data)} and \textit{Acrobot (large data)}, the dataset contains 1 million data samples and each method is only allowed to use the dataset once. We run all methods with small datasets as well. In \textit{Mountain Car, Cart Pole} and \textit{Acrobot}, the dataset contains 20000 data samples. In all experiments, the policy that is used to sample data is a policy that performs random actions in the environment. 

Table \ref{tab:control_results} shows control performances of all methods. We observe that gradient based methods outperform LSTD in experiments where the dataset is small. For example, batching SVRG and SVRG outperform LSTD in \textit{Cart Pole} and \textit{Acrobot}. In experiments where datasets are large, LSTD's performances improve. In \textit{Mountain Car (large data), Cart Pole (large data)} and \textit{Acrobot (large data)}, we run all gradient based methods for a single pass through the dataset. Since SVRG and SAGA need to compute full gradients at the beginning, they cannot solve the control tasks. GTD2, batching SVRG and SCSG do not rely on full gradients, so they make progress instantly. In particular, batching SVRG and SCSG achieve the same level of performance as LSTD in \textit{Cart Pole (large data)} and \textit{Acrobot (large data)}. Our proposed methods and LSTD both use the dataset once and solve the control tasks, while other gradient based methods need more than one pass of the data set. More importantly, unlike LSTD, our methods are first order methods and do not need to invert matrices, which makes our methods practical when both the size of the dataset and the number of features of the state are large.

\begin{figure}[hbt!]
    \center
    \includegraphics[width=0.5\linewidth]{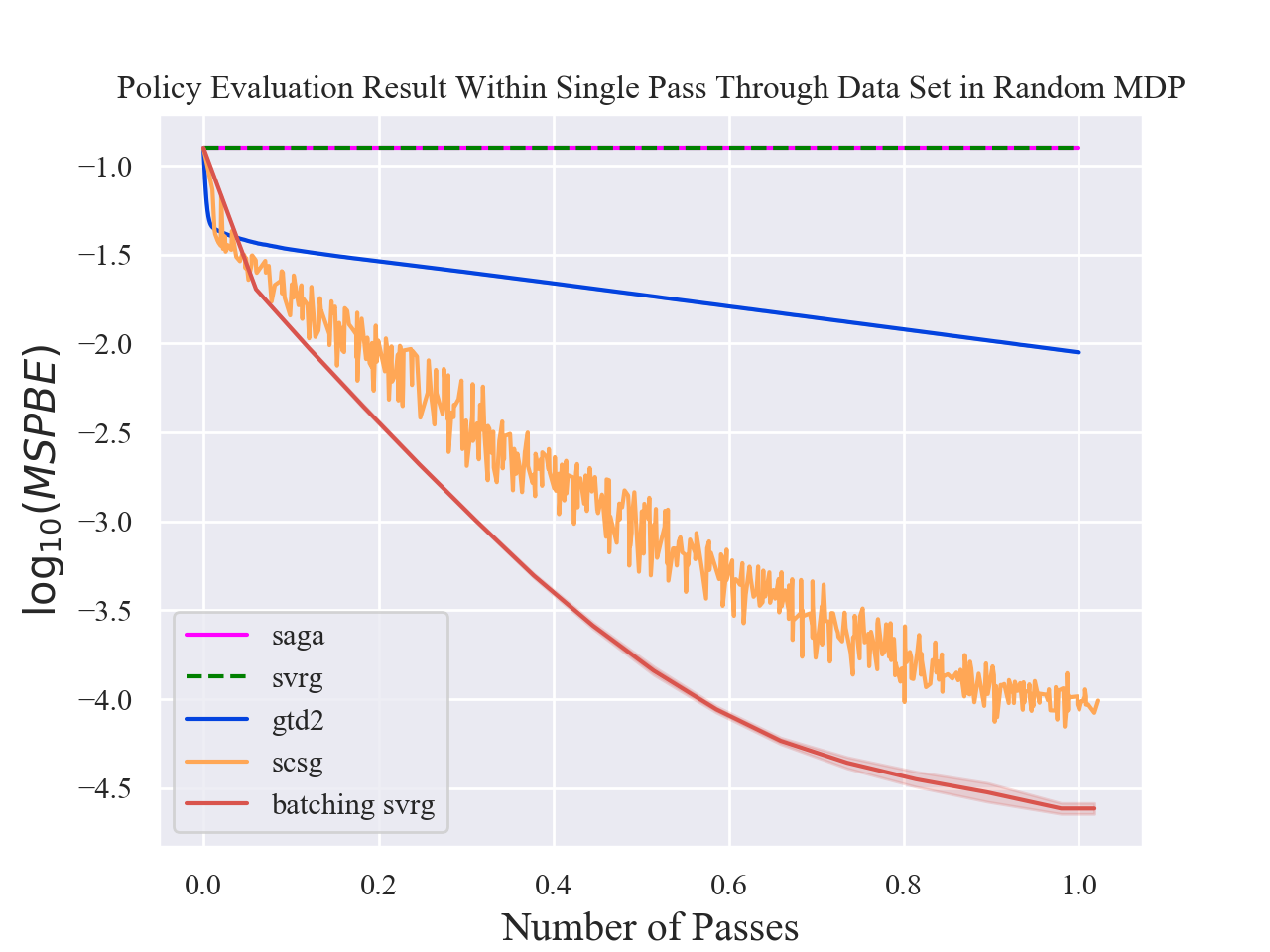}
    \caption{Policy evaluation performances in Random MDP environment within a single pass through the data set.}
    \label{fig:single_pass_rmdp}
\end{figure}

\subsection{Policy evaluation in large data settings}

To test our methods' performances on a very large dataset, we generate 10 million data samples from a policy that performs random actions in Random MDP. Figure \ref{fig:single_pass_rmdp} shows policy evaluation performances of all gradient based methods in Random MDP. We run all methods for a single pass through the dataset. SVRG and SAGA do not make progress because they need to compute full gradients. We observe that batching SVRG and SCSG converge much faster than GTD2, because batching SVRG and SCSG have linear convergence rates, while GTD2 has a sublinear convergence rate. This experiment shows again that our proposed methods have good performances in large data settings.

\section{CONCLUSION}

In this paper, we show that batching SVRG and SCSG converge linearly when solving the saddle-point formulation of MSPBE. This problem is convex-concave and is not strongly convex in the primal variable, so it is very different from the original objective function that batching SVRG and SCSG attempt to solve. Our algorithms are very practical because they require much fewer gradient evaluations than the vanilla SVRG for policy evaluation. There is a lot of room for applying more efficient optimization algorithms to problems in reinforcement learning, in order to obtain better theoretical guarantees and to improve sample and computational efficiency. We think the present work is a valuable contribution in that direction.

\bibliographystyle{apalike}
\bibliography{lib}

\appendix
\onecolumn
\section{PROOF OF THEOREM \ref{prop:one_epoch_anal}}
\label{appen:bsvrg_proof}

\begin{proof}
Define the residual vector $\Delta_m$ and $\Delta_{m,j}$ as:
\begin{equation} \label{def:delta_m_and_delta_mj}
\Delta_m = z_m - z^\star = 
          \left( \begin{matrix}
          \theta_m - \theta^* \\
          \frac{1}{\sqrt{\beta}}(\omega_m - \omega^*)
          \end{matrix} \right) \text{ and } \Delta_{m,j} = z_{m,j} - z^\star =
          \left( \begin{matrix}
          \theta_{m,j} - \theta^* \\
          \frac{1}{\sqrt{\beta}}(\omega_{m,j} - \omega^*)
          \end{matrix} \right)
\end{equation}

$\theta_m$ and $\omega_m$ are $\theta$ and $\omega$ at the beginning of epoch $m$. $\theta_{m,j}$ and $\omega_{m,j}$ are $\theta$ and $\omega$ at epoch $m$ and $j^{\text{th}}$ iteration of the inner loop . $\theta^*$ and $\omega^*$ are optimal solutions of (\ref{eq:saddle-point problem}). From the first order optimality condition, we know that

\begin{equation} \label{def:first_order_optimality}
        \left( \begin{matrix}
          0 & -\hat A^\top \\
          \hat A & \hat C
          \end{matrix}\right) 
          \left( \begin{matrix}
          \theta^* \\
          \omega^*
          \end{matrix} \right) = \left( \begin{matrix}
          0 \\
          \hat b
          \end{matrix} \right)
\end{equation}

The above equality is obtained by setting (\ref{eq:saddle-point gradient}) to a zero vector. 

By writing out algorithm \ref{alg:batch_svrg}'s update, we have: 

\begin{equation} \label{def:batch_svrg_update_z}
    z_{m,j+1} = z_{m,j} - \sigma_\theta \left(G_{t_j} z_{m,j} + (G_m - G_{t_j}) z_{m} - g_m \right)
\end{equation}
we defined $e_m$ the error coming from using a mini-batch to compute the gradients at epoch $m$.

\begin{equation} \label{def:batch_svrg_error_at_m}
    e_m = (G z_m - g) - (G_m z_m - g_m)
\end{equation}
we obtain then:
\begin{equation}
    z_{m,j+1} = z_{m,j} - \sigma_\theta \left(G_{t_j} z_{m,j} + (G - G_{t_j}) z_{m} - g - e_m \right)
\end{equation}
 Subtract both sides by $z^\star$ and use the first order optimality condition. We obtain:
           
\begin{align} \label{pf:bsvrg:delta_m_j+1}
    \Delta_{m, j+1} &= \Delta_{m,j} - \sigma_\theta (G\Delta_{m} + G_{t_j}\Delta_{m,j}-G_{t_j}\Delta_{m}) + \sigma_\theta e_m \nonumber \\
    &= (I-\sigma_\theta G)\Delta_{m,j} + \sigma_\theta (G-G_{t_j})(\Delta_{m,j}-\Delta_{m}) + \sigma_\theta e_m
\end{align}

We set $\beta =  \frac{8\lambda_{max}(\hat{A}^\top \hat{C}^{-1} \hat{A})}{\lambda_{min}(\hat{C})}$ so that $G$ is diagonalizable by Lemma \ref{lem:G_diag_cond}. Let $G = Q\Lambda Q^{-1}$ where $Q$ contains eigenvectors and $\Lambda$ contains eigenvalues of $G$. Multiply both sides of (\ref{pf:bsvrg:delta_m_j+1}) by $Q^{-1}$, then take squared 2-norm and expectation. Set $\delta = \Delta_{m,j}-\Delta_{m}$. We get:

\begin{align} \label{pf:pot_func_bd1}
    \enqds{\Delta_{m,j+1}} &= \EX \lVert Q^{-1}(I-\sigma_\theta G)\Delta_{m,j} + \sigma_\theta Q^{-1}(G-G_{t_j})\delta + \sigma_\theta Q^{-1} e_m \rVert^2 \nonumber\\
    &= \EX \norm{(I-\sigma_\theta \Lambda)Q^{-1}\Delta_{m,j}}^2 + \sigma_\theta^2 \EX \norm{ Q^{-1}(G-G_{t_j})\delta}^2 + 
    \sigma_\theta^2  \EX \norm{Q^{-1} e_m}^2 \nonumber \\
    &+ 2\sigma_\theta \EX \big<(I-\sigma_\theta \Lambda)Q^{-1}\Delta_{m,j}, Q^{-1}e_m \big> \nonumber\\
    &\leq \| I-\sigma_\theta \Lambda \|^2 \EX\norm{Q^{-1}\Delta_{m,j}}^2 + \sigma_\theta^2  \EX \norm{ Q^{-1}(G-G_{t_j})\delta}^2 + 
    \sigma_\theta^2 \EX \norm{Q^{-1}e_m}^2 \nonumber\\
    &+ 2\sigma_\theta \EX \big<(I-\sigma_\theta \Lambda)Q^{-1}\Delta_{m,j}, Q^{-1}e_m \big> \nonumber \\
    &\leq \| I-\sigma_\theta \Lambda \|^2 \EX\norm{Q^{-1}\Delta_{m,j}}^2 + \sigma_\theta^2  \EX \norm{ Q^{-1}G_{t_j}\delta}^2 + 
    \sigma_\theta^2 \EX \norm{Q^{-1}e_m}^2 \nonumber\\
    &+ 2\sigma_\theta \EX \big<(I-\sigma_\theta \Lambda)Q^{-1}\Delta_{m,j}, Q^{-1}e_m \big>
\end{align} 
The cross term in the second equality is simplified by using $\EX [G_{t_j}] = G$ and $G_{t_j}$ is independent with $\Delta_{m,j}$, $\Delta_{m}$ and $e_m$. We use in the last inequality that same independence and that the variance of a random variable is less than its second moment.

We borrow the following useful inequalities from appendix C of \cite{du:stoc_var_pe}.

\begin{equation} \label{eq:du_ineq1}
    \EX \norm{ Q^{-1}G_{t_j}\delta}^2  \leq 
    2 \kappa(Q)^2 L_G^2 (\enqds{\Delta_{m,j}} + \enqds{\Delta_{m}})
\end{equation}
    
\begin{equation} \label{eq:du_ineq2}
    \|  I - \sigma_\theta \Lambda \|^2 = \max\{ |1 - \alpha_k \lambda_{\min}(G)|^2, | 1 - \alpha_k \lambda_{\max}(G) |^2 \} \leq 1 - 2 \sigma_\theta \lambda_{\min} + \sigma_\theta^2 \kappa(Q)^2 L_G^2 \\
\end{equation}

Now we bound the cross term in (\ref{pf:pot_func_bd1}):

\begin{align}
    \label{pf:cross_term_bd}
    \EX \big<(I-\sigma_\theta \Lambda)Q^{-1}\Delta_{m,j}, Q^{-1}e_m \big> &\leq \sqrt{\EX \norm{(I-\sigma_\theta \Lambda)Q^{-1}\Delta_{m,j}}^2 } \sqrt{\EX\norm{Q^{-1}e_m}^2} \nonumber\\
    &\leq \frac{\lambda_{\min}}{4 \| I-\sigma_\theta\Lambda\|^2} \EX\norm{(I-\sigma_\theta \Lambda)Q^{-1}\Delta_{m,j}}^2 \nonumber\\
    &+ \frac{\| I-\sigma_\theta\Lambda\|^2}{ \lambda_{\min}}\EX \norm{Q^{-1}e_m}^2
\end{align}
the first inequality is obtained by Cauchy-Schwartz inequality
The last inequality follows from the fact that $2zw \leq a^{-1}z^2 + a w^2$ for any $a>0$ and we select $a = \frac{2 \| I-\sigma_\theta\Lambda\|^2}{ \lambda_{\min}}$ in order for the inequality to hold.

Put (\ref{eq:du_ineq1}), (\ref{eq:du_ineq2}) and (\ref{pf:cross_term_bd}) back to (\ref{pf:pot_func_bd1}). We obtain:

\begin{align} 
\enqds{\Delta_{m,j+1}} &\leq (1 - 2 \sigma_\theta\lambda_{\min} + 3 \sigma_\theta^2 k(Q)^2 L_G^2 + \frac{1}{2} \sigma_\theta \lambda_{\min} )\EX\norm{Q^{-1}\Delta_{m,j}}^2 \nonumber\\
& +  2\sigma_\theta^2\kappa^2(Q)L^2_G\EX\norm{Q^{-1}\Delta_{m}}^2  \nonumber\\
    &+ 2 \sigma_\theta \frac{1 - 2 \sigma_\theta\lambda_{\min} + \sigma_\theta^2 k(Q)^2 L_G^2 + (1/2) \sigma_\theta \lambda_{\min} }{ \lambda_{\min}}\EX \norm{Q^{-1}e_m}^2 
\end{align}

If we choose $0 \leq \sigma_\theta \leq \frac{\lambda_{\min}}{6 \kappa(Q)^2 L_G^2}$, then $3 \sigma_\theta^2 \kappa(Q)^2 L_G^2$ and $\sigma_\theta^2 \kappa(Q)^2 L_G^2$ are smaller than $(1/2) \sigma_\theta \lambda_{\min}$ which implies that: 

\begin{align} \label{pf:pot_func_bd2}
    \enqds{\Delta_{m,j+1}} &\leq (1 - \sigma_\theta\lambda_{\min} )\EX\norm{Q^{-1}\Delta_{m,j}}^2 \nonumber\\
& +  2\sigma_\theta^2\kappa^2(Q)L^2_G\EX\norm{Q^{-1}\Delta_{m}}^2  \nonumber\\
    &+ 2 \sigma_\theta \frac{1 - \sigma_\theta\lambda_{\min} }{ \lambda_{\min}} \EX \norm{Q^{-1}e_m}^2 
\end{align}

Note that $\sigma_\theta \lambda_{min} \leq \frac{\lambda_{min}^2}{6\kappa(Q)^2 L_G^2}\leq 1$, because of the following inequalities cited from Appendix C in \cite{du:stoc_var_pe}:

\begin{align} \label{pf:ineq_chain_lambda_klg}
    \lambda_{min}^2 \leq \lambda_{max}^2 \leq \norm{G}^2 = \norm{\EX G_t}^2 \leq \norm{\EX G_t^T G_t}^2 = L_G^2 \leq \kappa(Q)^2 L_G^2
\end{align}

Now enrolling the above inequality (\ref{pf:pot_func_bd2}) from $j=1$ to $K-1$, we obtain: 

\begin{align}
    \enqds{\Delta_{m+1}} &\leq (1 - \sigma_\theta\lambda_{\min} )^K \EX\norm{Q^{-1}\Delta_{m}}^2 \nonumber\\
& +  2\sigma_\theta^2\kappa^2(Q)L^2_G \sum_{j=0}^{K-1}  (1 - \sigma_\theta\lambda_{\min} )^j 
\EX\norm{Q^{-1}\Delta_{m}}^2  \nonumber\\
    &+ 2 \sigma_\theta \frac{1 - \sigma_\theta\lambda_{\min} }{ \lambda_{\min}} \sum_{j=0}^{K-1}  (1 - \sigma_\theta\lambda_{\min} )^j \EX \norm{Q^{-1}e_m}^2 
\end{align}

As 
\begin{equation}
\sum_{j=0}^{K-1}  (1 - \sigma_\theta\lambda_{\min} )^j = \frac{1 - (1-\sigma_\theta \lambda_{\min})^K}{1 - (1-\sigma_\theta \lambda_{\min})} \leq \frac{1}{\sigma_\theta \lambda_{\min} }
\end{equation}
, we obtain:
\begin{align}
    \enqds{\Delta_{m+1}} &  \leq \left( (1 - \sigma_\theta\lambda_{\min} )^K
+  \frac{2\sigma_\theta^2\kappa^2(Q)L^2_G}{\sigma_\theta \lambda_{\min}} \right) 
\EX\norm{Q^{-1}\Delta_{m}}^2  \nonumber\\
    &+ 2  \frac{1 - \sigma_\theta\lambda_{\min} }{ \lambda_{\min}^2} \EX \norm{Q^{-1}e_m}^2 
\end{align}

We choose:
\begin{equation}
    \sigma_\theta = \frac{\lambda_{\min}}{6 \kappa(Q)^2 L_G^2} \quad \text{and} \quad
    K = \frac{2}{\sigma_\theta \lambda_{\min}} = 
    \frac{12 \kappa(Q)^2 L_G^2}{\lambda_{\min}^2}
\end{equation}
then we get:

\begin{align} \label{eq: }
    \enqds{\Delta_{m+1}} &  \leq \left( \exp(-2) + 1/3 \right) 
\EX\norm{Q^{-1}\Delta_{m}}^2 + 2  \frac{1 - \sigma_\theta\lambda_{\min} }{ \lambda_{\min}^2} \EX \norm{Q^{-1}e_m}^2 \\
& \leq \frac{2}{3} \EX\norm{Q^{-1}\Delta_{m}}^2 + 
2  \frac{1 - \sigma_\theta\lambda_{\min} }{ \lambda_{\min}^2} \EX \norm{Q^{-1}e_m}^2
\end{align}

\end{proof}

\section{PROOF OF COROLLARY \ref{coroll: batch_svrg}}
\label{appendix: bsvrg_corollary}
\begin{proof}
By the definition of $e_m$ given in (\ref{eq: error def}), we have:

\begin{align} \label{pf:bsvrg_example_error_bd}
    \ens{e_m} &= \ens{G z_m - g - (G_m z_m - g_m)} \nonumber\\
    &= \ens{G z_m - g - \left(\frac{1}{B_m}\sum_{t \in \B_m} G_t z_m - g_t \right)}
\end{align}

Since $\sum_{t=1}^n G z_m - g - (G_t z_m - g_t) = n(G z_m - g) - n(G z_m -g) = 0$, we can apply lemma \ref{lem:e_helper} to (\ref{pf:bsvrg_example_error_bd}) and have:
\begin{align} \label{coroll:em_bound}
  &\ens{e_m} = \ens{G z_m - g - \left(\frac{1}{B_m}\sum_{t \in \B_m} G_t z_m - g_t \right)} \nonumber\\
  &= \frac{n-B_m}{n B_m}\frac{1}{n-1}\sum_{t=1}^{n} \norm{G z_m - g - (G_t z_m - g_t)}^2 \nonumber\\
  &= \frac{n-B_m}{n B_m}\frac{1}{n-1}\left[\underbrace{\sum_{t=1}^{n} \norm{G z_m - g}^2}_{n\norm{G z_m - g}^2} - 2\underbrace{\sum_{t=1}^{n}\langle G z_m - g \,, G_t z_m - g_t \rangle}_{n\norm{G z_m - g}^2} + \sum_{t=1}^{n} \norm{G_t z_m - g_t}^2\right] \nonumber\\
  &= \frac{n-B_m}{n B_m}\frac{1}{n-1}\left[\sum_{t=1}^{n} \norm{G_t z_m - g_t}^2 - \norm{G z_m - g}^2\right] \nonumber\\
  &\leq \frac{n-B_m}{n B_m}\Xi^2
\end{align}
The last inequality of the above derivation follows from the fact that we assume $\frac{1}{n-1} \sum_{t=1}^n \norm{G_t z - g_t}^2 - \norm{G z - g}^2 \leq \Xi^2 \quad \forall z$. Since we have set $B_m \geq \frac{n \Xi^2}{\Xi^2 + n \alpha \rho^m}$, by (\ref{coroll:em_bound}), we can bound $\ens{e_m}$ as 
\begin{equation}
    \EX \norm{e_m}^2 \leq \alpha \rho^m
\end{equation}

where $\alpha > 0$ and $\rho<2/3$. Combine the above inequality with the result in theorem \ref{prop:one_epoch_anal}. We obtain: 
\begin{align}
    \enqds{\Delta_{m+1} }\leq \frac{2}{3}\enqds{\Delta_m} + \frac{2(1 - \sigma_\theta\lambda_{\min})\norm{Q^{-1}}^2}{\lambda_{min}^2}\alpha\rho^m
\end{align}

Enroll the above inequality $M$ times. We have:

\begin{align}
    \enqds{\Delta_{m+1} }&\leq \left(\frac{2}{3}\right)^M\enqds{\Delta_0} + \frac{2\alpha(1 - \sigma_\theta\lambda_{\min})\norm{Q^{-1}}^2}{\lambda_{min}^2}\sum_{k=0}^{M-1}\frac{2}{3}^{M-1-k}\rho^k \nonumber\\
    &= \left(\frac{2}{3}\right)^M\enqds{\Delta_0} + \frac{2\alpha(1 - \sigma_\theta\lambda_{\min})\norm{Q^{-1}}^2}{\lambda_{min}^2}\left(\frac{2}{3}\right)^{M-1}\sum_{k=0}^{M-1}\left(\frac{3\rho}{2}\right)^k \nonumber\\
    &\leq \left(\frac{2}{3}\right)^M\enqds{\Delta_0} + \frac{2\alpha(1 - \sigma_\theta\lambda_{\min})\norm{Q^{-1}}^2}{\lambda_{min}^2}\left(\frac{2}{3}\right)^{M-1} \frac{1}{1-3\rho/2} \nonumber\\
    &= \left(\frac{2}{3}\right)^M\enqds{\Delta_0} + \frac{3\alpha(1 - \sigma_\theta\lambda_{\min})}{\lambda_{min}^2 (1-3\rho/2)}\norm{Q^{-1}}^2 \left(\frac{2}{3}\right)^M
\end{align}

The second inequality is derived by using the fact that $\frac{3\rho}{2} \leq 1$ because we set $\rho \leq \frac{2}{3}$.

\end{proof}

\section{PROOF OF THEOREM \ref{thm:scsg_result}}
\label{appendix: scsg_proof}
\begin{proof}
Algorithm \ref{alg:scsg}'s inner loop update is same with algorithm \ref{alg:batch_svrg}. Same as (\ref{def:batch_svrg_update_z}), we have: 
\begin{equation}
    z_{m,j+1} = z_{m,j} - \sigma_\theta \left(G_{t_j} z_{m,j} + (G_m - G_{t_j}) z_{m} - g_m \right)
\end{equation}

Define $e_m$ same as give in (\ref{def:batch_svrg_error_at_m}). We can write the above update equivalently as:

\begin{equation}
    z_{m,j+1} = z_{m,j} - \sigma_\theta \left(G_{t_j} z_{m,j} + (G - G_{t_j}) z_{m} - g - e_m \right)
\end{equation}

 Subtract both sides by $z^\star = (\theta^{\star}, \frac{1}{\sqrt{\beta}}\omega^{\star})^\top$ and use the first order optimality condition given in (\ref{def:first_order_optimality}). Define the residual vector $\Delta_m$ and $\Delta_{m,j}$ same as (\ref{def:delta_m_and_delta_mj}). We obtain:
           
\begin{align} \label{pf:scsg:delta_m_j+1}
    \Delta_{m, j+1} &= \Delta_{m,j} - \sigma_\theta (G\Delta_{m} + G_{t_j}\Delta_{m,j}-G_{t_j}\Delta_{m}) + \sigma_\theta e_m
\end{align}

By assumption \ref{assump:non-singularity} and our setting of $\beta$, conditions of Lemma \ref{lem:G_diag_cond} are satisfied, so $G$ is diagonalizable with all its eigenvalues real and positive. Let $G = Q\Lambda Q^{-1}$ where $Q$ contains eigenvectors and $\Lambda$ contains eigenvalues of $G$. Multiply both sides of (\ref{pf:scsg:delta_m_j+1}) by $Q^{-1}$, then take squared 2-norm and expectation. We get:

\begin{align} \label{pf:scsg:pot_func_bd1}
    \enqds{\Delta_{m_j+1}} &= \ens{Q^{-1}\Delta_{m,j} - \sigma_\theta (\Lambda Q^{-1}\Delta_m + Q^{-1}G_{t_j}\Delta_{m,j} - Q^{-1}G_{t_j}\Delta_m - Q^{-1}e_m)} \nonumber\\
    &= \enqds{\Delta_{m,j}} - 2\sigma_\theta \EX\left<\Lambda Q^{-1}\Delta_{m,j}-Q^{-1}e_m, Q^{-1}\Delta_{m,j} \right> \nonumber\\
    &+ \sigma_\theta^2 \underbrace{\ens{\Lambda Q^{-1}\Delta_m + Q^{-1}G_{t_j}\Delta_{m,j} - Q^{-1}G_{t_j}\Delta_m - Q^{-1}e_m}}_{\text{(1)}}
\end{align}

We first bound (1) in (\ref{pf:scsg:pot_func_bd1}). 

\begin{align} \label{pf:scsg:v_mj_bd}
&\ens{\Lambda Q^{-1}\Delta_m + Q^{-1}G_{t_j}\Delta_{m,j} - Q^{-1}G_{t_j}\Delta_m - Q^{-1}e_m} \nonumber\\
&= \ens{Q^{-1}G_{t_j}\Delta_{m,j} - Q^{-1}G_{t_j}\Delta_m - \Lambda Q^{-1}\Delta_{m,j} + \Lambda Q^{-1}\Delta_m} + \norm{\Lambda Q^{-1}\Delta_{m,j} - Q^{-1}e_m}^2 \nonumber\\
&= \enqds{(G-G_{t_j})(\Delta_{m,j}-\Delta_m)} + \norm{\Lambda Q^{-1}\Delta_{m,j} - Q^{-1}e_m}^2 \nonumber\\
&\leq \enqds{G_{t_j}(\Delta_{m,j}-\Delta_m)} + 2\norm{\Lambda Q^{-1}\Delta_{m,j}}^2 + 2\norm{Q^{-1}e_m}^2 \nonumber\\
&\leq 2 \kappa(Q)^2 L_G^2 \enqds{\Delta_{m,j}} + 2 \kappa(Q)^2 L_G^2\enqds{\Delta_{m}} + 2\norm{\Lambda Q^{-1}\Delta_{m,j}}^2 + 2\norm{Q^{-1}e_m}^2
\end{align}
To derive the first equality and the first inequality, we use the fact that $\ens{X} = \ens{X-\EX X} + \norm{\EX X}^2$ for any random variable $X$. We use (\ref{eq:du_ineq1}) to derive the last inequality. Put (\ref{pf:scsg:v_mj_bd}) back to (\ref{pf:scsg:pot_func_bd1}) and rearrange terms in (\ref{pf:scsg:pot_func_bd1}). We have:

\begin{align}
    2\sigma_\theta \EX \left<\Lambda Q^{-1}\Delta_{m,j}, Q^{-1}\Delta_{m,j}\right> &\leq \enqds{\Delta_{m,j}} - \enqds{\Delta_{m,j+1}} + 2\sigma_\theta \EX \left<Q^{-1}e_m, Q^{-1}\Delta_{m,j}\right> \nonumber\\
    &+ 2 \sigma_\theta^2\kappa(Q)^2 L_G^2 \enqds{\Delta_{m,j}} + 2\sigma_\theta^2 \kappa(Q)^2 L_G^2\enqds{\Delta_{m}} \nonumber\\
    &+ 2\sigma_\theta^2\norm{\Lambda Q^{-1}\Delta_{m,j}}^2 + 2\sigma_\theta^2\norm{Q^{-1}e_m}^2
\end{align}

Set $j$ to $K_m$ in the above inequality and take an expectation with respect to $K_m \sim \text{Geom}(\frac{B}{B+1})$, we obtain using Lemma \ref{lemma: geom_telescopic_sum}:
\begin{align}
    2\sigma_\theta \EX \left<\Lambda Q^{-1}\Delta_{m, K_m}, Q^{-1}\Delta_{m, K_m}\right> &\leq \frac{1}{B}\left(\enqds{\Delta_{m,0}} - \enqds{\Delta_{m, K_m}} \right) \nonumber\\
    &+ 2\sigma_\theta \EX \left<Q^{-1}e_m, Q^{-1}\Delta_{m, K_m} \right>+2 \sigma_\theta^2\kappa(Q)^2 L_G^2 \enqds{\Delta_{m,K_m}} \nonumber\\ 
    &+ 2\sigma_\theta^2 \kappa(Q)^2 L_G^2\enqds{\Delta_{m}}+ 2\sigma_\theta^2\EX\norm{\Lambda Q^{-1}\Delta_{m,K_m}}^2 + 2\sigma_\theta^2\norm{Q^{-1}e_m}^2
\end{align}

Multiply both sides of the above inequality with $B$. We get:
\begin{align} \label{pf:scsg:cross_term_bd}
    2\sigma_\theta B\EX \left<\Lambda Q^{-1}\Delta_{m+1}, Q^{-1}\Delta_{m+1}\right> &\leq \enqds{\Delta_{m}} - \enqds{\Delta_{m+1}} \nonumber\\
    &+ 2\sigma_\theta B\underbrace{\EX \left<Q^{-1}e_m, Q^{-1}\Delta_{m+1} \right>}_{\text{(1)}}+2B \sigma_\theta^2\kappa(Q)^2 L_G^2 \enqds{\Delta_{m+1}} \nonumber\\ 
    &+ 2B\sigma_\theta^2 \kappa(Q)^2 L_G^2\enqds{\Delta_{m}}+ 2B\sigma_\theta^2\underbrace{\EX\norm{\Lambda Q^{-1}\Delta_{m+1}}^2}_{\text{(2)}} \nonumber\\
    &+ 2B\sigma_\theta^2\norm{Q^{-1}e_m}^2
\end{align}

Note that $\Delta_{m,0} = \Delta_{m}$ and $\Delta_{m,K_m} = \Delta_{m+1}$ by definitions given in (\ref{def:delta_m_and_delta_mj}). We will derive an inequality and use it repeatedly in the rest of the proof:

\begin{align} \label{pf:scsg:iterate_to_func_bd}
    \enqds{\Delta_{m+1}} &= \ens{\Lambda^{-\frac{1}{2}}\Lambda^{\frac{1}{2}}Q^{-1}\Delta_{m+1} } \nonumber\\
    &\leq \norm{\Lambda^{-\frac{1}{2}}}^2\EX\norm{\Lambda^{\frac{1}{2}}Q^{-1}\Delta_{m+1}}^2 \nonumber\\
    &\leq \frac{1}{\lambda_{min}} \EX\Delta_{m+1}^\top Q^{-\top} \Lambda Q^{-1}\Delta_{m+1}
\end{align}

$\norm{\Lambda^{-\frac{1}{2}}}^2 \leq \frac{1}{\lambda_{min}}$ because $\Lambda$ is a diagonal matrix consisting of eigenvalues of $G$. We now give a bound of (1) in (\ref{pf:scsg:cross_term_bd})

\begin{align} \label{pf:scsg:cross_term_error_iterate_bd}
    \EX \left<Q^{-1}e_m, Q^{-1}\Delta_{m+1} \right> &\leq \sqrt{\enqds{e_m}} \sqrt{\enqds{\Delta_{m+1}}} \nonumber\\
    &\leq \frac{\lambda_{min}}{10}\enqds{\Delta_{m+1}} + \frac{10}{\lambda_{min}}\enqds{e_m} \nonumber\\
    &\leq 0.1\EX\Delta_{m+1}^\top Q^{-\top} \Lambda Q^{-1}\Delta_{m+1} + \frac{10}{\lambda_{min}}\enqds{e_m}
\end{align}

The first inequality follows from Cauchy-Schwartz inequality. The second inequaliy follows from the fact that $2zw \leq a^{-1}z^2 + a w^2$ for any $a>0$. The third inequality is derived by using (\ref{pf:scsg:iterate_to_func_bd}).

(2) in (\ref{pf:scsg:cross_term_bd}) can be bounded as:

\begin{align} \label{pf:scsg:grad_norm_squared_bd}
    \ens{\Lambda Q^{-1}\Delta_{m+1}} &= \EX\Delta_{m+1}^{\top}Q^{-\top}\Lambda^{2}Q^{-1}\Delta_{m+1} \nonumber\\
    &\leq \lambda_{max}\EX\Delta_{m+1}^{\top}Q^{-\top}\Lambda Q^{-1}\Delta_{m+1} \nonumber\\
    &= \lambda_{max} \EX \left<\Lambda Q^{-1}\Delta_{m+1}, Q^{-1}\Delta_{m+1}\right>
\end{align}




Put (\ref{pf:scsg:iterate_to_func_bd}), (\ref{pf:scsg:cross_term_error_iterate_bd}) and (\ref{pf:scsg:grad_norm_squared_bd}) back to (\ref{pf:scsg:cross_term_bd}), then rearrange terms. We have:

\begin{align}
    &\sigma_\theta B\left(1.8 - \frac{2\sigma_\theta \kappa(Q)^2 L_G^2}{\lambda_{min}} - 2\sigma_\theta \lambda_{max}\right)\EX\left<\Lambda Q^{-1}\Delta_{m+1}, Q^{-1}\Delta_{m+1}\right> + \enqds{\Delta_{m+1}} \nonumber\\
    &\leq \enqds{\Delta_m} + 2\sigma_\theta^2 B \kappa(Q)^2 L_G^2 \enqds{\Delta_m} + \sigma_\theta B\left(2\sigma_\theta+\frac{20}{\lambda_{min}} \right)\enqds{e_m}
\end{align}

We can lower bound the left hand side of the above inequality since $\frac{\lambda_{min}}{2}\enqds{\Delta_{m+1}} \leq \frac{1}{2}\Delta_{m+1}^\top Q^{-\top}\Lambda Q^{-1}\Delta_{m+1}$. Thus, we have:

\begin{align}
    &\sigma_\theta B\left(1.8 - \frac{2\sigma_\theta \kappa(Q)^2 L_G^2}{\lambda_{min}} - 2\sigma_\theta \lambda_{max}\right) \left[\frac{1}{2}\EX \Delta_{m+1}^\top Q^{-\top}\Lambda Q^{-1}\Delta_{m+1} + \frac{\lambda_{min}}{2}\enqds{\Delta_{m+1}}\right] \nonumber\\
    &+\enqds{\Delta_{m+1}} \leq \enqds{\Delta_m} + 2\sigma_\theta^2 B \kappa(Q)^2 L_G^2 \enqds{\Delta_m} + \sigma_\theta B\left(2\sigma_\theta+\frac{20}{\lambda_{min}} \right)\enqds{e_m}
\end{align}

From (\ref{pf:ineq_chain_lambda_klg}), we know that $- \frac{\kappa(Q)^2 L_G^2}{\lambda_{min}} \leq -\lambda_{max}$. We can further lower bound the left hand side of the above inequality:

\begin{align}
    &\sigma_\theta B\left(0.9 - \frac{2\sigma_\theta \kappa(Q)^2 L_G^2}{\lambda_{min}}\right) \left[\EX \Delta_{m+1}^\top Q^{-\top}\Lambda Q^{-1}\Delta_{m+1} + \lambda_{min}\enqds{\Delta_{m+1}}\right] 
    +\enqds{\Delta_{m+1}} \nonumber\\
    &\leq \enqds{\Delta_m} + 2\sigma_\theta^2 B \kappa(Q)^2 L_G^2 \enqds{\Delta_m} + \sigma_\theta B\left(2\sigma_\theta+\frac{20}{\lambda_{min}} \right)\enqds{e_m} \nonumber\\
    &\leq \enqds{\Delta_m} + 2\sigma_\theta^2 B \kappa(Q)^2 L_G^2 \enqds{\Delta_m} \nonumber\\
    &+ \sigma_\theta B\left(4\sigma_\theta+\frac{40}{\lambda_{min}} \right)\left[\frac{I(B < n)}{B} \kappa(Q)^2 L^2_G \EX \| Q^{-1}\Delta_m \|^2 
    + \frac{I(B < n)\| Q^{-1}\|^2}{B} \mathcal{H} \right]
\end{align}

In the second inequality, we use the result in Lemma \ref{lemma:error_m_bd} to bound $\enqds{e_m}$. We can use (\ref{pf:scsg:iterate_to_func_bd}) to upper bound $\enqds{\Delta_m}$ in the above inequality and we obtain:

\begin{align} \label{pf:scsg:lyapunov_prelim}
&\sigma_\theta B\left(0.9 - \frac{2\sigma_\theta \kappa(Q)^2 L_G^2}{\lambda_{min}}\right) \left[\EX \Delta_{m+1}^\top Q^{-\top}\Lambda Q^{-1}\Delta_{m+1} + \lambda_{min}\enqds{\Delta_{m+1}}\right]
    +\enqds{\Delta_{m+1}} \nonumber\\ 
&\leq \enqds{\Delta_m} + \sigma_\theta B \left(\frac{2\sigma_\theta \kappa(Q)^2 L^2_G}{\lambda_{min}} + \frac{I(B<n)\kappa(Q)^2 L^2_G}{B\lambda_{min}}\left(4\sigma_\theta+\frac{40}{\lambda_{min}}\right)\right)\EX \Delta_{m}^\top Q^{-\top}\Lambda Q^{-1}\Delta_{m} \nonumber\\
    &+ \sigma_\theta \left(4\sigma_\theta+\frac{40}{\lambda_{min}}\right)I(B<n)\norm{Q^{-1}}^2\mathcal{H}
\end{align}

By our setting, $\sigma_\theta \leq \frac{\lambda_{min}}{20\kappa(Q)^2 L_G^2}$, so $\frac{2\sigma_\theta \kappa(Q)^2 L_G^2}{\lambda_{min}}\leq \frac{1}{10}$. This implies that $0.9 - \frac{2\sigma_\theta \kappa(Q)^2 L_G^2}{\lambda_{min}} \geq 0.8$. We also know that $\lambda_{min}^2 \leq \kappa(Q)^2 L_G^2$ by (\ref{pf:ineq_chain_lambda_klg}), so $2\sigma_\theta \leq \frac{\lambda_{min}}{10\kappa(Q)^2 L_G^2} \leq \frac{1}{10\lambda_{min}} \leq \frac{1}{\lambda_{min}}$. Now we can lower bound and upper bound (\ref{pf:scsg:lyapunov_prelim}):

\begin{align} \label{pf:scsg:lyapunov_prelim2}
    &0.8\sigma_\theta B \EX \Delta_{m+1}^\top Q^{-\top}\Lambda Q^{-1}\Delta_{m+1} + \left(1+0.8\sigma_\theta B \lambda_{min} \right)\enqds{\Delta_{m+1}} \nonumber\\
    &\leq \enqds{\Delta_m} + \sigma_\theta B\left(0.1 + \frac{I(B<n)42\kappa(Q)^2 L_G^2 }{B \lambda_{min}^2} \right)\EX \Delta_{m}^\top Q^{-\top}\Lambda Q^{-1}\Delta_{m} \nonumber\\
    &+ \frac{I(B<n)42\sigma_\theta }{\lambda_{min}}\norm{Q^{-1}}^2\mathcal{H} \nonumber\\
    &\leq \enqds{\Delta_m} + 0.7\sigma_\theta B\EX \Delta_{m}^\top Q^{-\top}\Lambda Q^{-1}\Delta_{m} + \frac{I(B<n)42\sigma_\theta }{\lambda_{min}}\norm{Q^{-1}}^2\mathcal{H}
\end{align}

The last inequality of the above derivations follows from the fact that we set $B \geq \frac{70\kappa(Q)^2 L_G^2}{\lambda_{min}^2}$. 

Since we have set $\sigma_\theta \leq \frac{5}{28B\lambda_{max}} \leq \frac{5}{28B\lambda_{min}}$, we have $\frac{4}{5}\sigma_\theta B\lambda_{min} \leq \frac{1}{7}$. This implies that $1+0.8\sigma_\theta B\lambda_{min} \leq \frac{8}{7}$. We thus have $(1+0.8\sigma_\theta B\lambda_{min})\times 0.7 \leq \frac{8}{7} \times 0.7 = 0.8 $. Now we can lower bound the left hand side of (\ref{pf:scsg:lyapunov_prelim2}):

\begin{align} \label{pf:scsg:lyapunov_per_iter}
    &(1+0.8\sigma_\theta B\lambda_{min}) \left(\enqds{\Delta_{m+1}} + 0.7\sigma_\theta B \EX \Delta_{m+1}^\top Q^{-\top}\Lambda Q^{-1}\Delta_{m+1}\right) \nonumber\\
    &\leq \enqds{\Delta_m} + 0.7\sigma_\theta B\EX \Delta_{m}^\top Q^{-\top}\Lambda Q^{-1}\Delta_{m} + \frac{I(B<n)42\sigma_\theta }{\lambda_{min}}\norm{Q^{-1}}^2\mathcal{H}
\end{align}

Enrolling (\ref{pf:scsg:lyapunov_per_iter}) from $m=0$ to $M-1$, we have:

\begin{align}
    &(1+0.8\sigma_\theta B\lambda_{min})^M \left(\enqds{\Delta_{M-1}} + 0.7\sigma_\theta B \EX \Delta_{M-1}^\top Q^{-\top}\Lambda Q^{-1}\Delta_{M-1}\right) \nonumber\\
    &\leq \enqds{\Delta_0} + 0.7\sigma_\theta B\EX \Delta_{0}^\top Q^{-\top}\Lambda Q^{-1}\Delta_{0} + \frac{I(B<n)42\sigma_\theta }{\lambda_{min}}\norm{Q^{-1}}^2\mathcal{H}\sum_{j=0}^{M-1}(1+0.8\sigma_\theta B\lambda_{min})^j \nonumber\\
    &\leq (1+0.7\sigma_\theta B\lambda_{max})\enqds{\Delta_0} + \frac{I(B<n)42\sigma_\theta }{\lambda_{min}}\norm{Q^{-1}}^2\mathcal{H}\frac{(1+0.8\sigma_\theta B\lambda_{min})^M}{0.8\sigma_\theta B\lambda_{min}}
\end{align}

In the second inequality, we use the fact that $\Lambda$ is a diagonal matrix consisting of eigenvalues, which implies that: $\EX \Delta_{0}^\top Q^{-\top}\Lambda Q^{-1}\Delta_{0} \leq \lambda_{max}\enqds{\Delta_0}$. Divide both sides of the above inequality by $(1+0.8\sigma_\theta B\lambda_{min})^M$. We have:

\begin{align}
    &\enqds{\Delta_{M-1}} + 0.7\sigma_\theta B \EX \Delta_{M-1}^\top Q^{-\top}\Lambda Q^{-1}\Delta_{M-1} \nonumber\\
    &\leq \frac{(1+0.7\sigma_\theta B\lambda_{max})}{(1+0.8\sigma_\theta B\lambda_{min})^M}\enqds{\Delta_0}  + \frac{60\norm{Q^{-1}}^2 \mathcal{H} I(B<n)}{B\lambda_{min}^2}
\end{align}

Obviously, $\enqds{\Delta_{M-1}} \leq \enqds{\Delta_{M-1}} + 0.7\sigma_\theta B \EX \Delta_{M-1}^\top Q^{-\top}\Lambda Q^{-1}\Delta_{M-1}$, so we have:

\begin{align}
    \enqds{\Delta_{M-1}} \leq \frac{(1+0.7\sigma_\theta B\lambda_{max})}{(1+0.8\sigma_\theta B\lambda_{min})^M}\enqds{\Delta_0}  + \frac{60\norm{Q^{-1}}^2 \mathcal{H} I(B<n)}{B\lambda_{min}^2}
\end{align}

To conclude: 

\begin{align}
    &\EX\norm{\theta_{M-1} - \theta^\star}^2 \leq \ens{QQ^{-1}\Delta_{M-1}} \leq \norm{Q}^2\enqds{\Delta_{M-1}}\nonumber\\
    &\leq \frac{(1+0.7\sigma_\theta B\lambda_{max})\norm{Q}^2}{(1+0.8\sigma_\theta B\lambda_{min})^M}\enqds{\Delta_0}  + \frac{60\norm{Q}^2\norm{Q^{-1}}^2 \mathcal{H} I(B<n)}{B\lambda_{min}^2} \nonumber\\
    &\leq \frac{(1+0.7\sigma_\theta B\lambda_{max})\kappa(Q)^2}{(1+0.8\sigma_\theta B\lambda_{min})^M}\ens{\Delta_0}  + \frac{60\kappa(Q)^2 \mathcal{H} I(B<n)}{B\lambda_{min}^2}
\end{align}

\end{proof}

Here is the lemma that we restated from \citep{lei:scsg}. We use the same property of geometric distribution in the proof above.
\begin{lemma} \label{lemma: geom_telescopic_sum} Lemma A.2 in \citep{lei:scsg}.
Let $N \sim \text{Geom}(\gamma)$ for some $\gamma > 0$. Then any sequence $\{ D_n \}$ with $\EX | D_N | < \infty$,
\begin{equation}
    \EX[D_N - D_{N+1}] = \left( \frac{1}{\gamma} - 1 \right) (D_0 - \EX D_N)
\end{equation}
\end{lemma}

we need to show that $\mathbb{E} \| Q^{-1} \Delta_{m, K_m}\|^2< \infty$ for any $m$ where the expectation is with respect the geometric distributed variable $K_m$, so that we can apply lemma \ref{lemma: geom_telescopic_sum}.

\begin{lemma} \label{lemma: finite geom exp}
If $0 \leq \sigma_\theta \leq \frac{\lambda_{\min}}{6 \kappa(Q)^2 L_G^2}$, then for any $m$: 
\begin{equation}
    \mathbb{E} \| Q^{-1} \Delta_{m}\|^2 =  \mathbb{E} \| Q^{-1} \Delta_{m, K_m}\|^2< \infty
\end{equation}
\end{lemma}

\begin{proof}
We proceed by induction. When $m=0$, the statement is obvious. Suppose that the statement holds for $m$. i.e:
\begin{equation}
     \mathbb{E} \| Q^{-1} \Delta_{m}\|^2 =  \mathbb{E} \| Q^{-1} (z_m - z^\star)\|^2< \infty
\end{equation}
Let $y_{m, j}$ be sequence of iterates that would be produced by SVRG (by
computing the true gradient in the outer loop) initialized at $z_m = z_{m, 0}$ i.e: 
\begin{equation}
    y_m = y_{m, 0} = z_m, \quad y_{m,j+1} = y_{m,j} - \sigma_\theta \left(G_{t_j} y_{m,j} + (G - G_{t_j}) y_{m} - g \right)
\end{equation}
Recall the update of SCSG iterates $z_{m, j}$:
\begin{equation}
    z_{m,j+1} = z_{m,j} - \sigma_\theta \left(G_{t_j} z_{m,j} + (G - G_{t_j}) z_{m} - g - e_m \right)
\end{equation}
Then, we obtain:
\begin{equation}
    z_{m, j+1} - y_{m, j+1} = (I - \sigma_\theta G_{t_j}) (z_{m, j} - y_{m, j}) + \sigma_\theta e_m
\end{equation}
By left multiplying by $Q^{-1}$, taking the  square norm, expectation with respect to the mini-batch of outer loop ($\mathbb{E}[e_m] = 0$) and then expectation with respect to $t_j$ ($\mathbb{E}[G_{t_j}] = G$), we obtain:
\begin{align}
    \mathbb{E}\| Q^{-1}(z_{m, j+1} - y_{m, j+1})\|^2 & = \mathbb{E} \| Q^{-1}(I - \sigma_\theta G_{t_j}) (z_{m, j} - y_{m, j}) \|^2 + \sigma_\theta^2 \mathbb{E} \| Q^{-1}e_m\|^2 \\
    & = \| Q^{-1}(z_{m, j} - y_{m, j})\|^2 + \sigma_\theta^2 \mathbb{E} \| Q^{-1} G_{t_j} (z_{m, j} - y_{m, j}) \|^2 \nonumber 
    \\ 
    & \quad - 2 \sigma_\theta \mathbb{E}\langle Q^{-1}(z_{m, j} - y_{m, j}), Q^{-1} G_{t_j} (z_{m, j} - y_{m, j}) \rangle + \sigma_\theta^2 \mathbb{E} \| Q^{-1}e_m\|^2 \\
    & = \| Q^{-1}(z_{m, j} - y_{m, j})\|^2 + \sigma_\theta^2 \mathbb{E} \| Q^{-1} G_{t_j} (z_{m, j} - y_{m, j}) \|^2 \nonumber 
    \\ 
    & \quad - 2 \sigma_\theta \langle Q^{-1}(z_{m, j} - y_{m, j}), Q^{-1} G (z_{m, j} - y_{m, j}) \rangle + \sigma_\theta^2 \mathbb{E} \| Q^{-1}e_m\|^2 \\
    & = \| Q^{-1}(z_{m, j} - y_{m, j})\|^2 + \sigma_\theta^2 \mathbb{E} \| Q^{-1} G_{t_j} (z_{m, j} - y_{m, j}) \|^2 \nonumber 
    \\ 
    & \quad - 2 \sigma_\theta \langle Q^{-1}(z_{m, j} - y_{m, j}), \Lambda Q^{-1} (z_{m, j} - y_{m, j}) \rangle + \sigma_\theta^2 \mathbb{E} \|Q^{-1} e_m\|^2 \\
    & = \| Q^{-1}(z_{m, j} - y_{m, j})\|^2 + \sigma_\theta^2 \mathbb{E} \| Q^{-1} G_{t_j} (z_{m, j} - y_{m, j}) \|^2 \nonumber 
    \\ 
    & \quad - 2 \sigma_\theta \| \Lambda^{1/2} Q^{-1}(z_{m, j} - y_{m, j}) \|^2 + \sigma_\theta^2 \mathbb{E} \| Q^{-1}e_m\|^2 \\
    & \leq \| Q^{-1}(z_{m, j} - y_{m, j})\|^2 + 
    \sigma_\theta^2 \kappa(Q)^2 L_G^2 \| Q^{-1}(z_{m, j} - y_{m, j})\|^2 \nonumber \\
    & \quad -2 \sigma_\theta \lambda_{\min} \| Q^{-1}(z_{m, j} - y_{m, j})\|^2 + \sigma_\theta^2 \mathbb{E} \| Q^{-1}e_m\|^2
\end{align}
where we use in the last inequality the fact that $\mathbb{E} \| Q^{-1} G_{t_j} (z_{m, j} - y_{m, j}) \|^2 \leq \kappa(Q)^2 L_G^2 \| Q^{-1}(z_{m, j} - y_{m, j})\|^2$. As $0 \leq \sigma_\theta \leq \frac{\lambda_{\min}}{6 \kappa(Q)^2 L_G^2}$, we obtain: 

\begin{equation}
     \mathbb{E}\| Q^{-1}(z_{m, j+1} - y_{m, j+1})\|^2 \leq 
     (1 - \sigma_\theta \lambda_{\min}) \| Q^{-1}(z_{m, j} - y_{m, j})\|^2 + \sigma_\theta^2 \mathbb{E} \| e_m\|^2
\end{equation}
By enrolling the last inequality $j$ times, we obtain:

\begin{align} \label{eq: z-y}
    \mathbb{E}\| Q^{-1}(z_{m, j} - y_{m, j})\|^2 & \leq 
     (1 - \sigma_\theta \lambda_{\min})^{j} \| Q^{-1}(z_{m, 0} - y_{m, 0})\|^2 + j \sigma_\theta^2 \mathbb{E} \| Q^{-1}e_m\|^2 \nonumber \\
     & = j \sigma_\theta^2 \mathbb{E} \| Q^{-1}e_m\|^2
\end{align}
Now we need to bound $\mathbb{E}\|Q^{-1}(y_{m, j} - z^{\star})\|^2$, As $y_{m, j}$ differ from batching SVRG iterates only by the error term $e_m$, we can use the inequality (\ref{pf:pot_func_bd2}) and set $e_m =0$ to obtain:

\begin{align}
    \enqds{(y_{m, j+1} - z^{\star})} &\leq (1 - \sigma_\theta\lambda_{\min} )\EX\norm{Q^{-1}(y_{m, j} - z^{\star})}^2 +  2\sigma_\theta^2\kappa^2(Q)L^2_G\EX\norm{Q^{-1}(y_{m} - z^{\star})}^2 \\
    & \leq \EX\norm{Q^{-1}(y_{m, j} - z^{\star})}^2 +  2\sigma_\theta^2\kappa^2(Q)L^2_G\EX\norm{Q^{-1}(y_{m} - z^{\star})}^2
\end{align}
By enrolling the inequality above $j$ times, we obtain:

\begin{align} \label{eq: y - z star}
    \enqds{(y_{m, j} - z^{\star})} & \leq \EX\norm{Q^{-1}(y_{m, 0} - z^{\star})}^2 + 2j \sigma_\theta^2\kappa^2(Q)L^2_G\EX\norm{Q^{-1}(y_{m} - z^{\star})}^2 \nonumber \\
    & = (1 + 2j \sigma_\theta^2\kappa^2(Q)L^2_G ) \EX\norm{Q^{-1}(y_{m} - z^{\star})}^2 \nonumber \\
    & \leq ( 1 + j  \sigma_\theta \lambda_{\min}) \EX\norm{Q^{-1}(z_{m} - z^{\star})}^2 
\end{align}
Putting (\ref{eq: z-y}) and (\ref{eq: y - z star}) together, we obtain:

\begin{align} \label{eq: decomp error}
    \enqds{\Delta_{m, j}} & \leq 2 \enqds{(z_{m, j} - y_{m, j})} + 2 \enqds{(y_{m, j} - z^{\star})} \nonumber \\
    & \leq 2 j \sigma_\theta^2 \mathbb{E} \| Q^{-1}e_m\|^2 + 2 ( 1 + j  \sigma_\theta \lambda_{\min}) \EX\norm{Q^{-1}\Delta_m}^2 
\end{align}

By the inequality (\ref{eq: error bound}) in proof of lemma \ref{lemma:error_m_bd}, we have:
\begin{equation}
    \mathbb{E} \| Q^{-1}e_m\|^2 \leq 2\kappa(Q)^2 L^2_G \EX \| Q^{-1}\Delta_m \|^2 
    + 2\| Q^{-1}\|^2 \mathcal{H}
\end{equation}

By (\ref{eq: decomp error}),
\begin{align}
    \enqds{\Delta_{m, j}} & \leq 4 j \sigma_\theta^2 \kappa(Q)^2 L^2_G \EX \| Q^{-1}\Delta_m \|^2 + 4 j \sigma_\theta^2 \| Q^{-1}\|^2 \mathcal{H}
    + 2 ( 1 + j  \sigma_\theta \lambda_{\min}) \EX\norm{Q^{-1}\Delta_m}^2 \nonumber \\
    & \leq 4 j \sigma_\theta \lambda_{\min} \EX \| Q^{-1}\Delta_m \|^2 + 4 j \sigma_\theta^2 \| Q^{-1}\|^2 \mathcal{H}
    + 2 ( 1 + j  \sigma_\theta \lambda_{\min}) \EX\norm{Q^{-1}\Delta_m}^2 \nonumber \\
    & = 2 j \sigma_\theta (3 \lambda_{\min} \EX\norm{Q^{-1}\Delta_m}^2 + 2\sigma_\theta \| Q^{-1}\|^2 \mathcal{H}) + 2 \EX\norm{Q^{-1}\Delta_m}^2
\end{align}

By setting $j=K_m$ and taking expectation with respect to the number of inner loop iteration $K_m$ and applying the induction hypothesis, we obtain

\begin{equation}
    \enqds{\Delta_{m+1}} \leq 2 \EX[K_m] \sigma_\theta (3 \lambda_{\min} \EX\norm{Q^{-1}\Delta_m}^2 + 2\sigma_\theta \| Q^{-1}\|^2 \mathcal{H}) + 2 \EX\norm{Q^{-1}\Delta_m}^2 < \infty
\end{equation}
\end{proof}

In the proof of theorem \ref{thm:scsg_result} and lemma \ref{lemma: finite geom exp}, we use a bound on the error term $\enqds{e_m}$. Now we present the proof of this bound.

\begin{lemma} \label{lemma:error_m_bd}
\begin{align*}
    \enqds{e_m} \leq \frac{2I(B < n)}{B} \kappa(Q)^2 L^2_G \EX \| Q^{-1}\Delta_m \|^2 
    + \frac{2I(B < n)\| Q^{-1}\|^2}{B} \mathcal{H}
\end{align*}
\end{lemma}

\begin{proof}
\begin{align}
 \EX \norm{Q^{-1}e_m}^2 & = 
 \EX \norm{Q^{-1}e_m + Q^{-1}(G_m z^\star-g_m) - Q^{-1}(G_m z^\star-g_m) }^2 \nonumber \\
 & \leq 2 \EX \norm{Q^{-1}e_m + Q^{-1}(G_m z^\star-g_m)}^2 + 2 
 \EX \norm{Q^{-1}(G_m z^\star-g_m)}^2 \nonumber \\
 & = 2 \EX \norm{Q^{-1}e_m + Q^{-1}(G_m z^\star-g_m) - \underbrace{(G z^\star -g)}_{=0}}^2 + 2 
 \EX \norm{Q^{-1}(G_m z^\star-g_m)}^2 \nonumber \\
 & = 2\EX \norm{Q^{-1} (G_m - G) \Delta_m}^2
 + 2\EX \norm{Q^{-1}(G_m z^\star-g_m)}^2 \nonumber \\ 
 & = 2\EX \EX_{\mathcal{B}} \norm{\frac{1}{B} \sum_{t \in \mathcal{B}} Q^{-1} (G_t - G) \Delta_m}^2 
 + 2\EX \EX_{\mathcal{B}} \norm{\frac{1}{B} \sum_{t \in \mathcal{B}}Q^{-1}(G_t z^\star-g_t)}^2
\end{align}
Where $\EX$ is expectation over $\B$.
As
\begin{equation}
    \frac{1}{n} \sum_{t=1}^n Q^{-1} (G_t - G) \Delta_m = Q^{-1} (G - G) \Delta_m = 0 \quad \text{and} \quad 
    \frac{1}{n} \sum_{t=1}^n Q^{-1}(G_t z^\star-g_t) = G z^\star - g = 0
\end{equation}
then, by applying lemma \ref{lem:e_helper}, we obtain

\begin{align}\label{eq: error bound}
    \EX \norm{Q^{-1}e_m}^2 & \leq 
    \frac{2I(B < n)}{B} \EX \frac{1}{n} \sum_{t=1}^n \| Q^{-1}(G_t - G) \Delta_m \|^2 
    + \frac{2I(B < n)\| Q^{-1}\|^2}{B}  \underbrace{\frac{1}{n} \sum_{t=1}^n \| G_t z^\star - g_t \|^2}_{=\mathcal{H}} \nonumber \\
    & = \frac{2I(B < n)}{B} \EX \frac{1}{n} \sum_{t=1}^n \| Q^{-1}(G_t - G) \Delta_m \|^2 
    + \frac{2I(B < n)\| Q^{-1}\|^2}{B} \mathcal{H} \nonumber \\
    & \leq  \frac{2I(B < n)}{B} \EX \frac{1}{n} \sum_{t=1}^n  \| Q^{-1} G_t \Delta_m \|^2 
    + \frac{2I(B < n)\| Q^{-1}\|^2}{B} \mathcal{H} \nonumber \\
    & \leq \frac{2I(B < n)}{B} \kappa(Q)^2 L^2_G \EX \| Q^{-1}\Delta_m \|^2 
    + \frac{2I(B < n)\| Q^{-1}\|^2}{B} \mathcal{H}
\end{align}
We used in the third inequality the fact that variance of random variable is upper bounded by its second moment. We used the last inquality the fact that 
$ \frac{1}{n} \sum_{t=1}^n \| Q^{-1} G_t \Delta_m \|^2 \leq \kappa(Q)^2 L^2_G \EX \| Q^{-1}\Delta_m \|^2 $ as shown in \cite{du:stoc_var_pe}.
\end{proof}

\begin{lemma} \label{lem:e_helper}
Restated from lemma B.1 in \citep{lei:scsg}: Let $z_1 ... z_M \in {\rm I\!R}^d$ be an arbitrary population of $M$ vectors with $\sum_{j=1}^M z_j = 0$. Let $\mathcal{I}$ be a uniform random subsets of $[M]$ with size $m$. Then,
\begin{align*}
    \ens{\frac{1}{m}\sum_{i\in \mathcal{I}}z_i} = \frac{M-m}{(M-1)m}\times\frac{1}{M}\sum_{j=1}^M \norm{z_j}^2 \leq \frac{I(m < M)}{m} \frac{1}{M}\sum_{j=1}^M \norm{z_j}^2
\end{align*}
where $I(m < M) = 1$ if $m < M$ and zero otherwise
\end{lemma}

\section{PROOF OF COROLLARY \ref{cor:scsg_comp_cost}}
\label{appendix: scsg_corollary}
\begin{proof}
From theorem \ref{thm:scsg_result}, we know that:
\begin{align}
\EX\norm{\theta_{M-1} - \theta^\star}^2 \leq \frac{(1+0.7\sigma_\theta B\lambda_{max})\kappa(Q)^2}{(1+0.8\sigma_\theta B\lambda_{min})^M}\ens{\Delta_0}  + \frac{60\kappa(Q)^2 \mathcal{H} I(B<n)}{B\lambda_{min}^2}
\end{align}

If $\max\left\{\frac{70\kappa(Q)^2 L_G^2}{\lambda_{min}^2}, \frac{120\kappa(Q)^2\mathcal{H}}{\lambda_{min}^2 \epsilon}\right\} < n$, then $B \geq \frac{120\kappa(Q)^2\mathcal{H}}{\lambda_{min}^2 \epsilon}$, implying that $\frac{60\kappa(Q)^2\mathcal{H}}{B\lambda_{min}^2} \leq \frac{\epsilon}{2}$. We can thus write the above inequality equivalently as:

\begin{align} \label{pf:scsg:convergence_result_prelim}
\EX\norm{\theta_{M-1} - \theta^\star}^2 \leq \frac{(1+0.7\sigma_\theta B\lambda_{max})\kappa(Q)^2}{(1+0.8\sigma_\theta B\lambda_{min})^M}\ens{\Delta_0} + \frac{\epsilon}{2}
\end{align}

In the case where $\max\left\{\frac{70\kappa(Q)^2 L_G^2}{\lambda_{min}^2}, \frac{120\kappa(Q)^2\mathcal{H}}{\lambda_{min}^2 \epsilon}\right\} \geq n$, $\frac{60\kappa(Q)^2 \mathcal{H} I(B<n)}{B\lambda_{min}^2} = 0$, so the above inequality still holds.

By our choice of $\sigma_\theta$, $\sigma_\theta \leq \frac{5}{28 B \lambda_{max}}$, so $\sigma_\theta B \lambda_{max} \leq \frac{5}{28}$. Right hand side of (\ref{pf:scsg:convergence_result_prelim}) can be bounded as:

\begin{align}
\EX\norm{\theta_{M-1} - \theta^\star}^2 \leq \frac{1.125\times\kappa(Q)^2}{(1+0.8\sigma_\theta B\lambda_{min})^M}\ens{\Delta_0} + \frac{\epsilon}{2}    
\end{align}



$\EX\norm{\theta_{M-1} - \theta^\star}^2 \leq \epsilon$ if $M \geq \frac{\log\left(\frac{2.25\kappa(Q)^2\ens{\Delta_0}}{\epsilon} \right)}{\log\left(1+0.8\sigma_\theta B\lambda_{min}\right)}$, so the number of outer loop iterations that Algorithm \ref{alg:scsg} needs to take in order to reach an $\epsilon$-optimal solution is:
\begin{align*}
 \frac{\log\left(\frac{2.25\kappa(Q)^2\ens{\Delta_0}}{\epsilon} \right)}{\log\left(1+0.8\sigma_\theta B\lambda_{min}\right)} 
 &= O\left(\left(1 + \frac{1}{\sigma_\theta B\lambda_{min}} \right)\log\left(\frac{\kappa(Q)^2\ens{\Delta_0}}{\epsilon} \right)\right) \\
 &= O\left(\left(1 + \frac{\kappa(Q)^2 L_G^2}{B\lambda_{min}^2} \right)\log\left(\frac{\kappa(Q)^2\ens{\Delta_0}}{\epsilon} \right)\right)
\end{align*}


The second equality follows from $\sigma_\theta B \lambda_{min} = \min\left\{\frac{\lambda_{min}}{20\kappa(Q)^2 L_G^2}, \frac{5}{28B\lambda_{max}} \right\}B\lambda_{min}$. If $\frac{B\lambda_{min}^2}{20\kappa(Q)^2 L_G^2} < \frac{5\lambda_{min}}{28\lambda_{max}} < 1$, $\frac{1}{\sigma_\theta B \lambda_{min}} = O\left( \frac{\kappa(Q)^2 L_G^2}{B\lambda_{min}^2}\right)$. Otherwise, $\frac{1}{\sigma_\theta B \lambda_{min}} = O(1)$.

In each iteration of the outer loop in Algorithm \ref{alg:scsg}, a batch of $B$ gradients are computed before entering the inner loop, then, the inner loop executes for $O(B)$ iterations in expectations. In total, $O(B)$ gradient evaluations are required in expectations during each epoch of Algorithm \ref{alg:scsg}. The total computation cost in expectation is:

\begin{align*}
   &O\left(\left(B + \frac{\kappa(Q)^2 L_G^2}{\lambda_{min}^2} \right)d\times\log\left(\frac{\kappa(Q)^2\ens{\Delta_0}}{\epsilon} \right)\right) \\
   &= O\left(\left(\min\left\{\max\left\{\frac{70\kappa(Q)^2 L_G^2}{\lambda_{min}^2}, \frac{120\kappa(Q)^2\mathcal{H}}{\lambda_{min}^2 \epsilon}\right\}, n\right\} + \frac{\kappa(Q)^2 L_G^2}{\lambda_{min}^2} \right)d\times \log\left(\frac{\kappa(Q)^2\ens{\Delta_0}}{\epsilon} \right)\right) \\
   &= O\left(\left(\min\left\{ \frac{\kappa(Q)^2\mathcal{H}}{\lambda_{min}^2 \epsilon}, n\right\} + \frac{\kappa(Q)^2 L_G^2}{\lambda_{min}^2} \right)d\times \log\left(\frac{\kappa(Q)^2\ens{\Delta_0}}{\epsilon} \right)\right)
\end{align*}

\end{proof}

\section{EXPERIMENT DETAILS}
\label{appen:exp_detail}
\begin{itemize}
\item \textbf{Environments}:
\begin{enumerate}
\item Random MDP environment \citep{dann:survey}. A randomly generated MDP with 400 states, 10 actions. Each state has a 201 dimensional feature vector where each entry except the last one is uniformly sampled from $[0, 1]$. The last entry of every state's feature vector is set to 1. Transition probabilities and rewards are uniformly sampled from $[0, 1]$. Discount factor is 0.95.
\item We used Open AI Gym's MountainCar-v0, CartPole-v1 and Acrobot-v1 \citep{openai_gym} to simulate mountain car, cart pole and acrobot environments. Discount factor is set to 0.99 in mountain car and cart pole experiments. In acrobot experiments, discount factor is 0.9. We changed the reward function in cart pole experiments. Agent receives a reward of -1 when it fails and 0 otherwise.
\end{enumerate} 
\item \textbf{Data collection process}. Data is generated by evaluating a policy that performs random actions in the environment.
\item \textbf{Parameter selection}. The number of SVRG's inner loop iterations is searched from $n\times1$ and $n\times2$, where $n$ is the size of the dataset. Step sizes are searched from $\{10, 1, 10^{-1}, \dots, 10^{-6} \}$. We perform a grid search for these parameters on a validation dataset in each environment.
\item \textbf{Feature engineering process}
\begin{enumerate}
    \item In random mdp experiments, we do not modify state's feature vector.
    \item In mountain car experiments, we normalize state's features to $[0, 1]$ and apply them to 10 evenly spaced RBF kernels, defined as $exp(-r/\sigma)$ where $\sigma=0.01$.
    \item In cart pole experiments, we apply state's features to 3 RBF kernels. For car's position values and pole's angle values, means of RBF kernels are -0.1, 0, 0.1. For velocity values of the car and pole, means of RBF kernels are -1, 0, 1. $\sigma=0.5$.
    \item In acrobot experiments, we normalize state's features to $[0, 1]$ and apply them to 3 evenly spaced RBF kernels. We only use cosine and sine of two links' angles as our features. $\sigma=0.1$.
\end{enumerate}
For experiments that use RBF kernels, we normalize feature vectors so they sum to 1. Emprical matrices $\hat{A}$ and $\hat{C}$ are not full rank in mountain car, cart pole and acrobot experiments, we add a small identity matrix ($10^{-5}\times I$) to $\hat{A}$, $\hat{A_t}$, $\hat{C}$, $\hat{C_t}$ when running gradient based methods and LSTD.

\item \textbf{Description of how experiments were run}.
\begin{enumerate}
    \item In the experiment that compares batching SVRG and SVRG, We sample a dataset which contains 5000 data samples, then run each method for 10 times and aggregate results. Initial values of $\theta$ and $\omega$ are zero vectors, except in acrobot experiments, where entries of $\theta$ and $\omega$ are randomly sampled from $[0, 10]$, because zero vectors give a small objective value.
    \item When comparing control performances, we first sample a dataset from a policy that performs random actions, then we run gradient based methods for 100 passes through the dataset in small data settings and 1 pass in large data settings; finally we report the number of steps it takes the agent to terminate by executing the learned policy. In small data settings, this process is repeated for 50 times. In large data settings, this process is repeated for 20 times. Initial values of $\theta$ and $\omega$ are zero vectors.
    \item In the policy evaluation experiment under large data setting of random MDP, we sample a dataset with 10 million data samples and run each method for a single pass through the dataset. We report results from 10 runs. Initial values of $\theta$ and $\omega$ are zero vectors.
\end{enumerate}

\item \textbf{Computing infrastructure} All experiments are performed on a Linux compute node by using 6 cores, 16 gigabytes of RAM and a single GPU. 
\end{itemize}
\end{document}